\documentclass[lettersize,journal]{IEEEtran}
\usepackage{amsmath,amsfonts}
\usepackage{algorithmic}
\usepackage{algorithm}
\usepackage{array}
\usepackage{multirow}
\usepackage[caption=false,font=normalsize,labelfont=sf,textfont=sf]{subfig}
\usepackage{textcomp}
\usepackage{stfloats}
\usepackage{url}
\usepackage{verbatim}
\usepackage{graphicx}
\usepackage{cite}
\usepackage{hyperref}
\begin{document}

\title{Model Compression Techniques in Biometrics Applications: A Survey}
  
  

\author{\IEEEauthorblockN{Eduarda Caldeira\IEEEauthorrefmark{1}\IEEEauthorrefmark{2}\IEEEauthorrefmark{5}, Pedro C. Neto\IEEEauthorrefmark{1}\IEEEauthorrefmark{2}, Marco Huber, Naser Damer, Ana F. Sequeira\IEEEauthorrefmark{1}\IEEEauthorrefmark{2}}

    \IEEEauthorblockA{\IEEEauthorrefmark{1}INESC TEC}\\
    \IEEEauthorblockA{\IEEEauthorrefmark{2}FEUP, University of Porto}\\
    \IEEEauthorblockA{\IEEEauthorrefmark{3}Fraunhofer Institute for Computer Graphics Research IGD}\\
    \IEEEauthorblockA{\IEEEauthorrefmark{4}TU Darmstadt}\\
    \IEEEauthorblockA{\IEEEauthorrefmark{5}Corresponding author: up201906930@edu.fe.up.pt}
}

\markboth{}
{}
\maketitle

\begin{abstract}
The development of deep learning algorithms has extensively empowered humanity's task automatization capacity. However, the huge improvement in the performance of these models is highly correlated with their increasing level of complexity, limiting their usefulness in human-oriented applications, which are usually deployed in resource-constrained devices. This led to the development of compression techniques that drastically reduce the computational and memory costs of deep learning models without significant performance degradation. This paper aims to systematize the current literature on this topic by presenting a comprehensive survey of model compression techniques in biometrics applications, namely quantization, knowledge distillation and pruning. We conduct a critical analysis of the comparative value of these techniques, focusing on their advantages and disadvantages and presenting suggestions for future work directions that can potentially improve the current methods. Additionally, we discuss and analyze the link between model bias and model compression, highlighting the need to direct compression research toward model fairness in future works.


\textit{Impact Statement}---This work constitutes, to the extent of the authors’ knowledge, the first review addressing compression techniques in biometric applications. Compression is particularly important in this type of application due to the deployment of biometrics algorithms in edge devices with limited resource availability to be used in real-time applications. Furthermore, biometrics is prone to bias, whose effects are particularly harmful in algorithms based on human data, as they can result in discrimination against underrepresented demographic groups. This is also addressed in detail in this survey, bridging the previously existing gap in the literature and showing that there is still a big opportunity for evolution in this area, as some hidden implications of compression are still not completely understood, resulting in an undesired drop in model fairness.

\end{abstract}


\section{Introduction}\label{sec:intro}

Deep learning (DL) models have been widely studied in the last decade, resulting in a wide variety of tools that can perform extremely complex tasks in several different areas \cite{neto2022imil4path,melo2023synthesis}. Although the development of DL algorithms has extensively empowered humanity's task automatization capacity, the huge improvement in the performance of these models is highly correlated with their increasing level of complexity, which is associated with problems that cannot be disregarded, such as high memory demands and increased computational cost \cite{kolf2022lightweight,vitek2023ipad}. Black-box behavior is also associated with complexity, resulting in models that lack interpretability \cite{neto2022explainable}. Biometrics applications are no exception. In industry/organization-oriented scenarios, such as biometrics-based surveillance systems, access to cutting-edge technology is common and, thus, the negative impact of model complexity is reduced. However, automated technologies have already reached an evolutionary state that allows them to bring benefits to humans as individuals. In theory, the algorithms used by the industry could perform similar tasks in human-oriented frameworks. In practice, however, this is not possible because human-oriented applications are mainly intended to be used in resource-constrained devices, such as mobile phones \cite{delgado2023m,kocacinar2022real} or head-mounted devices \cite{boutros2020benchmarking,DBLP:journals/ivc/BoutrosDRRKK20,miller2022temporal}, meaning that their design should take additional factors and challenges into consideration \cite{gou2021knowledge,ge2018low}. On one side, the application will be handled by a device with very limited computational power and memory availability, meaning that it should be optimized for low resource availability and power consumption ~\cite{ge2018low,krishnamoorthi2018quantizing}. On the other side, different from what happens while using devices such as computers, humans are not used to waiting when using mobile devices, expecting their applications to run almost instantly. None of these criteria is met by DL models, posing severe limitations to their usage in human-oriented applications \cite{gou2021knowledge,gholami2022survey}. Hence, the progress associated with complex models is not directly useful to systems that need to be deployed in embedded devices to address the users' needs, such as face recognition (FR) systems. 

This question has gathered DL researchers' attention, resulting in attempts to develop lighter neural networks for human-centered tasks. According to Zhu~\textit{et al.}~\cite{zhu2017prune}, simpler models can be obtained either by training a simpler architecture from scratch or by compressing a heavy model with state-of-the-art (SOTA) performance. Although training models with fewer parameters from scratch has the potential to reduce the computational burden of a specific task, the ability to model complex functions is smaller in a simpler architecture. While a more complex architecture would learn how to interpret relevant features to the task at hand, ensuring that a smaller network trained from scratch is focusing on the important information is not trivial \cite{wang2021teacher,zhu2017prune}. To the extent of our knowledge, this technique has not proved to reach competitive results yet \cite{wang2021teacher,neto2023compressed} and, thus, deviates from the scope of this study. Compression techniques, on the other hand, result in models that are based on complex networks, which gives rise to a different interpretation: instead of only extracting information from the data, compressed models also filter the relevant knowledge previously learned by their complex versions, which eases their learning process. These techniques have already proved to drastically reduce the computational and memory costs of SOTA DL models without significant performance degradation, proving to be useful tools in developing applications with limitations imposed by the low availability of computational resources, namely in biometrics applications \cite{boutros2022quantface,ge2018low,vitek2023ipad}.  

Model compression techniques can be divided into three main groups \cite{choi2020data}: \textbf{quantization}, \textbf{knowledge distillation} (KD) and \textbf{pruning}. Despite the importance of these techniques, their study has been mainly focused on computer vision (CV) tasks \cite{zhou2017incremental,jacob2018quantization,li2016pruning,zhu2017prune}, while much less attention has been given to compressing biometrics algorithms \cite{boutros2022quantface,luo2016face,polyak2015channel}. Furthermore, while compression techniques can give rise to outstanding results, it is crucial to study them in detail. However, compression's performance in the literature is mainly linked to the overall performance of the resulting algorithm, without any special concern regarding biases or specific sub-groups on the original data. This flawed overall analysis might contribute to the hidden behaviors of these deep systems, leading to the prevalence of problems intrinsically connected with compression and undetected due to an improper evaluation procedure. As an example, the model might be losing evaluation capabilities in classes that are underrepresented in the training data, resulting in increased bias \cite{neto2023compressed}, which is undesired. This reveals the need to conduct extensive studies on compression-induced bias. The discrimination of underrepresented groups is particularly harmful when dealing with data directly extracted from humans, as happens in biometrics applications.

Based on the outline above, we present this survey that covers the compression works done in biometrics. In scenarios where research on compression is still underdeveloped in this field, we also leveraged the wider literature on computer vision. We start by mathematically defining the compression methods quantization, KD and pruning (Section \ref{sec:compression}). Then, we explore the current literature on compression to propose a few research directions for researchers working on this topic (Section \ref{sec:sota}). An overview of the analyzed studies can be found on our GitHub page~\footnote{\href{https://github.com/EduardaCaldeira/compression_bias_survey}{GitHub}}.
We also addressed the problem of compression-induced bias by highlighting relevant works in this area, as well as the contributions, points of improvement and future interesting research directions of each one of them (Section \ref{sec:adv_vs_disadv}). Finally, we summarize the performed analysis while highlighting our main findings (Section \ref{sec:conclusion}). Hence, this work presents relevant contributions regarding model compression techniques, namely: 

\begin{enumerate}
    \item a thorough analysis of different compression techniques, from both conceptual and mathematics perspectives;
    \item a systematization of the literature dedicated to quantization, knowledge distillation and pruning, highlighting biometrics works;
    \item a critical analysis of the existent compression techniques, highlighting their advantages and disadvantages, while presenting suggestions of future work directions that can potentially improve the current methods;
    \item a detailed discussion on model bias, in general, and compression-induced bias, in particular, highlighting the need to consider model fairness in future compression research.
\end{enumerate}



\section{Compression}\label{sec:compression}

This section presents preliminaries covering the understanding of different compression techniques as well as their advantages and 
drawbacks. 
\subsection{Quantization}

Most DL models developed nowadays are full-precision (FP) models that represent their weights and activations as 32-bit parameters. These models have achieved remarkable performances in several tasks in the past decades \cite{neto2022myope,neto2022imil4path,melo2023synthesis}. However, FP operations are more computationally expensive than low-precision (LP) operations. This means that the model can be accelerated during training and inference by using LP representation \cite{boutros2022quantface}, enabling its deployment in novel scenarios, namely real-time applications. Besides, FP representation is also expensive in terms of memory consumption, resulting in models that cannot be deployed in resource-constrained devices such as mobile phones \cite{boutros2022pocketnet}. 

\begin{figure}
    \centering
    \includegraphics[scale=0.45]{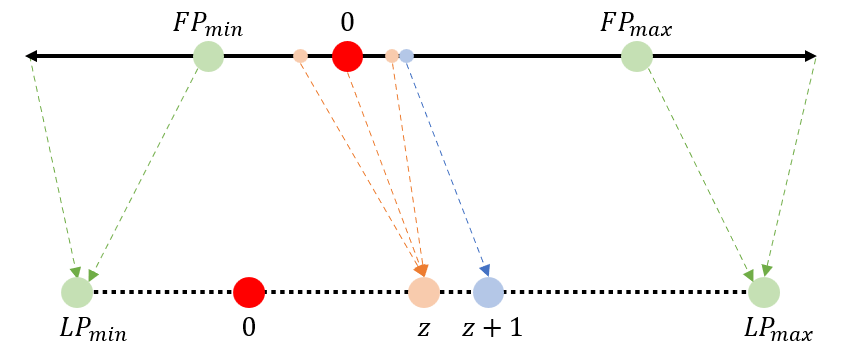}
    \caption{Quantization framework of a signed uniform affine quantizer. The top and bottom axes represent the continuous FP and the discrete quantized spectrums, respectively. The orange and green arrows represent the effects of the rounding and clipping functions (Equations \ref{eq:quant1} and \ref{eq:quant2}), respectively.}
    \vspace{-1.5em}
    \label{fig:quant}
\end{figure}

One way to compress FP models is model quantization. Model quantization is the process of converting an FP model into a possibly faster and lighter model by reducing the precision with which its parameters are represented to a lower number of bits (Figure~\ref{fig:quant}). Although several studies have been conducted to analyze 8-bit quantization with outstanding results \cite{jacob2018quantization,wang2022learnable}, any parameter precision can be used including the most extreme case, 1-bit quantization \cite{gholami2022survey}, where each parameter can only take one of two possible values. These lower bit-precision models have also been able to achieve near FP accuracy, especially when large networks are being considered \cite{krishnamoorthi2018quantizing}. This may be attributed to the fact that over-parametrized models can be altered in several different ways and still achieve optimal performance \cite{gholami2022survey}.

Quantized models maintain the same architecture as their FP versions, which makes quantization protocols much easier to design and implement than other compression techniques, such as KD, which will be discussed in Section \ref{sec:KD_SOTA}. Furthermore, the representation of the model's weights and/or activations as LP integers supports even faster inference and computation since several hardware platforms, libraries and processors allow for faster processing and inference of 8-bit data \cite{krishnamoorthi2018quantizing}. For instance, compared to its FP version, Pythorch \cite{pytorch2019} can reduce a quantized model's running time and memory bandwidth by a factor of two to four \cite{boutros2022quantface, krishnamoorthi2018quantizing, pytorch2019}.

Four different factors, discussed in this section, need to be considered when performing quantization: the quantization function (uniform, non-uniform, learnable), the quantization strategy (WOQ, AOQ, WAQ), the granularity (GW, LW, CW) and the global algorithm (PTQ, QAT), among other detailed hyper-parameters. 

\subsubsection{Quantization Function}
The most straightforward way to perform quantization is to define a \textbf{uniform quantizer} that evaluates the FP parameters and, given the final number of bits, $b$, defines a set of quantization thresholds. Consider that we want to represent a set of $k$ parameters $p_i$, $i\in\{1, ..., k\}$ as integers with $b$-bit precision. The number of different values that each quantized parameter, $q_i$, can take is $N = 2^b$. To convert the FP values into LP ones, two parameters have to be determined: the \textbf{scale}, $s$, which depends both on $N$ and on the range of values assumed by $p_i$ and the \textbf{zero-point}, $z$, to which the FP zero will map to. The FP range can be simply obtained by considering the lowest and highest values assumed by $p_i$, $p_{min}$ and $p_{max}$. Using a smaller range of values is possible but eliminates the quantizer's capacity to distinguish between parameters within the truncated range while enhancing its discriminating ability for the remaining ones. Although using a wider range of values might seem unnecessarily harmful to the discriminative ability of the quantizer, it might be needed in some situations since the FP range should be relaxed to include the FP zero. This way, operations that use the FP zero (such as zero-padding) will not lead to quantization errors \cite{krishnamoorthi2018quantizing}. Hence, the FP range, $(FP_{min}, \; FP_{max})$, should be defined as:\vspace{-1em}

\begin{equation}
    (FP_{min}, \;FP_{max})=
    \begin{cases}
    (0, \;p_{max}),\quad if\; p_{min}, \;p_{max} > 0\\
    (p_{min}, \;0),\quad if\; p_{min}, \;p_{max} < 0\\
    (p_{min}, \;p_{max}),\quad otherwise
    \end{cases}.
    \label{eq:FP_range}
\end{equation}

$FP_{min}$ and $FP_{max}$ are then used to determine the value of $s$ (Equation \ref{eq:scale}) and perform the quantization (Equation \ref{eq:quant1}): \vspace{-1em}

\begin{equation}
    s=\frac{FP_{max}-FP_{min}}{LP_{max}-LP_{min}},
    \label{eq:scale}
    \vspace{-1em}
\end{equation}

\begin{equation}
    p_{int_i}=\varphi(\frac{p_i}{s})+z,
    \label{eq:quant1}
\end{equation}
where $\varphi(.)$ is the selected rounding function. A clipping function is then applied to clip the values outside the low-precision range to its limits:\vspace{-1em}

\begin{equation}
    p_{quant_i}=
     \begin{cases}
    LP_{min},\quad if\; p_{int_i} < LP_{min} \\
    LP_{max},\quad if\; p_{int_i} > LP_{max} \\
    p_{int_i},\quad otherwise
    \end{cases},
    \label{eq:quant2}
\end{equation}
where $(LP_{min},\; LP_{max})$ define the low precision range; these values correspond to $(0, \;N-1)$ or $(-N/2, \;N/2-1)$ for unsigned and signed quantization, respectively. The dequantization process can be defined as:\vspace{-1em}

\begin{equation}
    p_{float_i}=(p_{quant_i}-z)\times s.
    \label{eq:dequant}
\end{equation}

Equations \ref{eq:quant1} and \ref{eq:quant2} define a \textbf{uniform affine quantizer}, which can be further simplified to a \textbf{uniform symmetric quantizer} by considering that the FP zero maps to 0 \cite{gholami2022survey}. In this case, $z=0$, and Equations~\ref{eq:quant1} and~\ref{eq:dequant} can be rewritten as:\vspace{-1em}

\begin{equation}
    p_{int_i}=\varphi(\frac{p_i}{s}),
    \label{eq:quant1_sym}
    \vspace{-1em}
\end{equation}

\begin{equation}
    p_{float_i}=p_{quant_i}\times s.
    \label{eq:dequant_sym}
\end{equation}

Another way to quantize a set of parameters is to select a \textbf{non-uniform quantization procedure} based on a predefined function, such as a \textbf{logarithmic quantizer} \cite{miyashita2016convolutional}, which works as a uniform quantizer in the logarithmic domain. However, these pre-defined quantizers might not result in optimal quantization due to the complexity of DL models. Hence, learnable quantization methods have been proposed \cite{zhang2018lq,jeon2022mr,wang2022learnable}. These methods minimize the error induced by quantization by learning how to interpret and adapt the models to which it is applied. Although some \textbf{learnable quantization} methods have already been developed, this area of quantization is still underdeveloped and should be further studied, since it has proved to be beneficial to the final performance of the model \cite{zhang2018lq,jeon2022mr,wang2022learnable}.

\subsubsection{Quantization Strategy}
Despite the importance of selecting an appropriate quantization strategy, it is also extremely relevant to choose where to apply it, since both weights and activations can be quantized. \textbf{Weight-only quantization} (WOQ) is a straightforward computation since the range of the FP weights is known in advance, leading to a deterministic quantization that can be directly performed. \textbf{Activation-only quantization} (AOW), however, requires the usage of a calibration dataset to determine the activation range when non-saturating functions like ReLU \cite{nair2010rectified} are being used. When non-saturating activations are used, the quantization result depends on the calibration dataset, which makes AOW non-deterministic \cite{krishnamoorthi2018quantizing} and leads to less precise quantization than when the values of the activation function lie within a fixed range \cite{jacob2018quantization}. \textbf{Weight-activation quantization} (WAQ) consists of applying WOQ and AOQ simultaneously. 

\subsubsection{Quantization Granularity}
There is also the need to group the parameters that will be quantized together and, thus, define each FP range. Three types of strategies can be considered \cite{kolf2022lightweight,krishnamoorthi2018quantizing}:

\begin{itemize}
    \item \textbf{group-wise (GW) quantization}: some layers of the model are grouped. Each group has its own FP range;
    \item \textbf{layer-wise (LW) quantization}: each layer defines a FP range and is quantized accordingly. For layers with more than one filter, all the channels are considered when determining the FP range \cite{bunda2022sub,jacob2018quantization,zhou2017incremental};
    \item \textbf{channel-wise (CW) quantization}: each filter is independently quantized with its own FP range \cite{kolf2022lightweight,boutros2022quantface,zhang2018lq}.
\end{itemize}

Selecting the parameters that define a certain FP range with finer granularity (CW quantization) means that more quantization computations will be performed. However, this extra computational cost does not have any impact on the model's inference time. Besides, the selection of a finer granularity might lead to better results, since batch normalization can cause huge variations across the dynamic range of each layer's filters, leading to higher FP ranges with less discriminative power when performing LW or GW quantization \cite{krishnamoorthi2018quantizing}.

\subsubsection{Global Algorithm}
The quantization procedure can be applied in two distinct ways: \textbf{post-training quantization} (PTQ) and \textbf{quantization-aware training} (QAT). PTQ consists of pre-training the FP network that will be quantized and applying one of the previously defined quantization strategies afterward. This is done because training quantized models from scratch usually leads to worse performances than starting with a pretrained network that can transfer part of its knowledge to its quantized version (similarly to what happens for KD \cite{wang2021teacher,hinton2015distilling,aslam2023privileged} and pruning \cite{yu2018nisp,zhu2017prune}). However, quantization generally leads to a drop in performance that grows larger as $b$ decreases. To reduce this gap and regain some of the lost performance, a fine-tuning step can be added after the PTQ is complete. This step is known as QAT and makes use of simulated quantizations to compute the gradients during the training procedure. In these cases, the forward and the backward propagation steps can be performed with either FP \cite{kolf2022lightweight,jacob2018quantization} or simulated quantized \cite{krishnamoorthi2018quantizing} weights and activations. Either way, the gradients that are used to update the model's FP parameters are calculated considering their quantized version. At the end of each training iteration, the gradient updates the model's FP parameters which are then quantized to allow the next gradient computation to be performed. It should be noted that the gradients of the quantized parameters are most likely zero, leading to an ineffective update of the model's parameters during QAT. To avoid this problem, the value of the gradient should be approximated by a pre-defined function. A commonly used example is the Straight-Through Estimator (STE), which approximates the gradient of the quantized parameters to the gradient of the identity function \cite{kolf2022lightweight,krishnamoorthi2018quantizing}.

Although QAT may seem hard to handle at first, the procedure itself can be simple. In general, QAT allows to improve the performance achieved with simple PTQ. Nonetheless, the fact that QAT requires retraining the model should not be disregarded, especially when the access to training data is compromised due to privacy reasons \cite{jeon2022mr}.

\subsection{Knowledge Distillation}
\label{sec:KD_intro}

\begin{figure}
    \centering
    \includegraphics[scale=0.37]{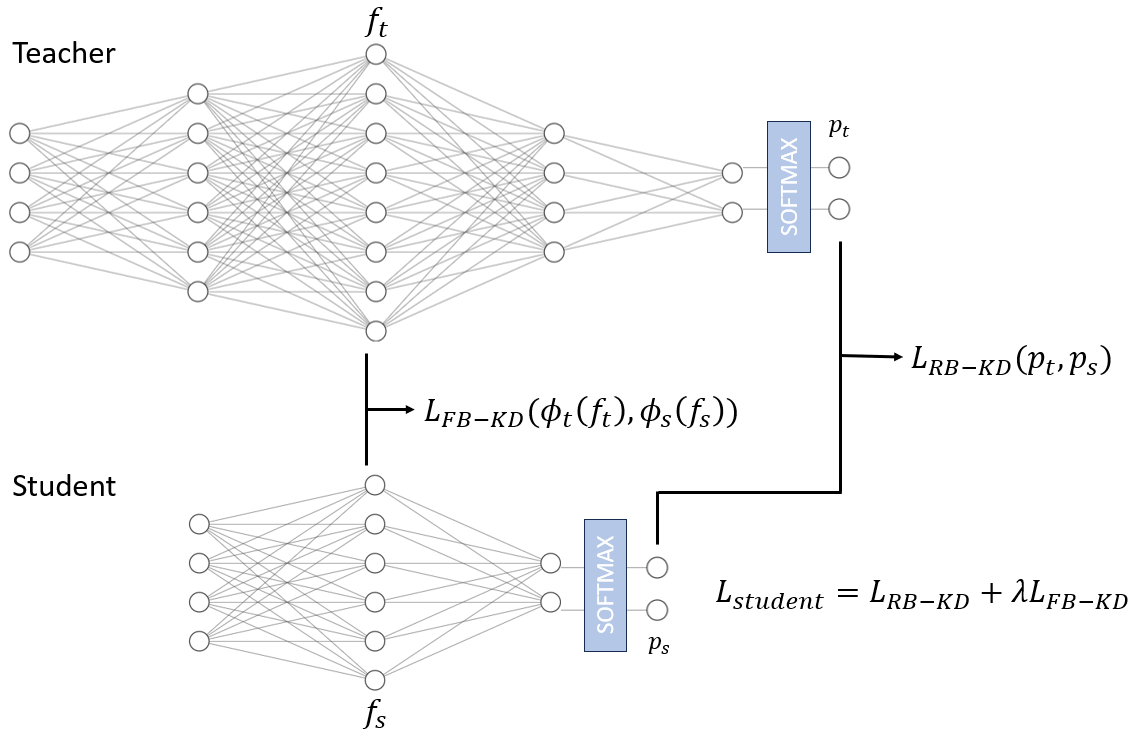}
    \caption{Knowledge distillation framework consisting of RB-KD and FB-KD terms ($L_{RB-KD}$ and $L_{FB-KD}$, respectively). These terms are weighted by the hyperparameter $\lambda$, resulting in the student's loss, $L_{student}$.}
    \vspace{-1.5em}
    \label{fig:KD}
\end{figure}

The general idea behind KD is very familiar to humans as this technique is based on our learning process \cite{gou2021knowledge}. When someone starts studying a new subject, they need to find the appropriate tools to learn. Although these tools can come from several sources, they are usually impersonated by someone who already has knowledge on the topic: a teacher. In this case, the student's learning process is guided by the teacher, who acts as an external agent who transfers their knowledge on the topic to the student. In KD applications, the idea is similar. The teacher is a complex model that performs well in the proposed task, while the student is a model that faces a knowledge gap. This gap can be due to low data availability, creating challenging scenarios where the student has to analyze harder ~\cite{huber2021mask} or lower-quality \cite{ge2018low,ge2020efficient,boutros2022low} data. However, KD is also often used to achieve compression \cite{gou2021knowledge}, as illustrated in Figure~\ref{fig:KD}. In this case, the knowledge gap arises from the lower complexity of the student, meaning that the teacher learns the proposed task more easily. Hence, distilling the teacher's knowledge to the student model may improve its performance, avoiding misinterpretations of the provided data by redirecting the student toward a path known to produce good results. Therefore, KD can be used to train lightweight student models that mimic the behavior of more complex teacher models, providing compact solutions with acceptable performance degradation~\cite{gou2021knowledge}.

\subsubsection{KD Strategy}
KD can be applied in several stages of the learning process. Supervised learning is a way of distilling knowledge itself since it provides the model with feedback about its predictions by comparing them to ground truth (GT) information provided by an expert labeler. In this situation, there is not a model that works as the teacher, but knowledge is distilled from this expert, who works as a teacher to the student model. \textbf{Response-based} KD (RB-KD) \cite{ge2020efficient,zhao2023grouped} algorithms follow a similar approach, by taking a step back and considering the soft probabilities produced by the teacher network as the source of knowledge distilled to the student. That way, the student model can directly learn the predictions of the teacher \cite{gou2021knowledge}. This type of KD is applied based on the assumption that SOTA models may produce predictions that constitute a better representation of reality than the training data labels since some information about the relation between the input data and the different classes is lost when the final prediction is made. However, it should be noted that RB-KD has severe limitations. First, these methodologies rely solely on the model's final predictions, which reduces their usage to supervised approaches \cite{gou2021knowledge}. Second, the fact that the distillation is performed at the probabilities level implies that the data used in the KD step is the same used to train the teacher. This is not always possible, especially in the biometrics field, as it works with sensitive human-based information \cite{kolf2023syper}. Using a distinct dataset to perform the KD step usually results in reduced performance \cite{kolf2023syper} without solving the privacy concern, as labeled data is still needed. Besides, despite relaxing some of the constraints imposed by the usage of GT labels, directly deriving the probabilities to transfer knowledge through the usage of the softmax function over the teacher's logits generally results in probability vectors where several entries are collapsed to near 0 values \cite{wang2021knowledge}, resulting in lost information. This problem can be addressed by the usage of a temperature parameter in the softmax function, as explained later in this section. However, even when dealing with supervised approaches where RB-KD can be used, its impact on the final performance of the student can be small, since the information related to the teacher's intermediate layers is not accessed by the student \cite{gou2021knowledge}. Keeping this information in a black-box severely restrains the amount of useful knowledge that is distilled to the student, which is particularly limiting when the teacher's architecture is very deep and, thus, hard to understand without proper insight \cite{gou2021knowledge}. 

\textbf{Feature-based} KD (FB-KD) \cite{luo2016face,duong2019shrinkteanet,wu2020learning,ge2020efficient,wang2021teacher,huber2021mask,liu2022coupleface,boutros2022low,li2023rethinking} helps mitigate the problem of ignoring useful teacher knowledge by performing the knowledge transfer of one or more intermediate layers of the teacher to the student, providing a much deeper insight into the teacher's thought process. The usage of this type of methodology allows the student to easily identify the parameters that need to be improved to better mimic the teacher's behavior, which favors its tuning process. Besides, the usage of FB-KD also addresses the data privacy issue associated with RB-KD strategies, as transferring knowledge on the feature level does not require the teacher and the student to be trained on the same data or even labeled data \cite{kolf2023syper}. This allows for the easy usage of privacy-friendly unlabeled data, namely synthetic data \cite{kolf2023syper}. Although FB-KD is more flexible since the knowledge can be distilled from several different combinations of intermediate layers, its higher complexity poses some challenges. First, this higher flexibility raises questions regarding which and how many layers should be chosen to better distill the teacher's knowledge to the student. The answer to this type of question is not trivial, especially when there is a big complexity gap between the teacher and the student \cite{aslam2023privileged}. If the architectures of the two models are very different, it may be impossible to directly distill the teacher's knowledge from intermediate layers, meaning that the KD step can only be performed through the usage of an algorithm that translates the teacher's knowledge to the student \cite{duong2019shrinkteanet} (reducing the features' dimensions, for example). Such a translator is not only hard to find but also increases the architecture's computational cost \cite{zhao2023grouped}, which poses a big disadvantage.

\textbf{Relation} KD (R-KD) \cite{duong2019shrinkteanet,liu2022coupleface,boutros2022template,huang2022evaluation,caldeira2023unveiling,aslam2023privileged} aims to solve the problem of directly transferring knowledge from the teacher's feature space to the student's. Instead of rigorously mimicking the teacher's space as happens when FB-KD is used, R-KD softens the knowledge transfer process by distilling relational information contained in this space. This is usually done through the usage of similarity metrics such as cosine similarity (CS) \cite{duong2019shrinkteanet,liu2022coupleface,caldeira2023unveiling}. This way, instead of strictly trying to preserve the architecture of the teacher's feature space, which might be impossible especially when there is a large knowledge gap between the teacher and the student \cite{aslam2023privileged}, only the structural relations contained within this space are distilled, resulting in a simplified process with the potential to boost KD results.

The KD step is usually accounted for by changes in the loss function used to update the model. Although these changes are conditioned by the type of KD that is being applied, they usually allow for the usage of several loss metrics to compare the teacher and student networks. In RB-KD algorithms, the model's loss, $L_{model}$ can be simply rewritten as:\vspace{-1em}

\begin{equation}
    p_{m,i} = \frac{e^{z_i}}{\sum_{j=1}^{N}{e^{z_j}}}, m \in {t, s}
    \label{eq:logits_noT},
\end{equation}

\begin{equation}
    L_{model} = L_{KD}(p_t, p_s),
\end{equation}
where $z_t$ and $z_s$ are the logits produced by the teacher and the student, respectively, $p_t$ and $p_s$ are the probabilities that the teacher and the student assigned to each class, respectively, $p_{m,i}$ is the $i$-th entry of vector $p_m$ (which might take the value of $p_t$ or $p_s$), $N$ is the number of classes and $L_{KD}(.)$ is the selected KD loss function. It should be noted that this type of KD sometimes does not consider a specific term for the classification loss, since the considered probabilities already contain information that accounts for that loss. However, a classification term can also be considered to include the GT information encoded by the samples' labels: \vspace{-1em}

\begin{equation}
    L_{model} = L_{class}(c_{GT}, c_s) + \lambda L_{KD}(p_t, p_s),
    \label{eq:rKD_class}
\end{equation}
where $c_{GT}$ and $c_s$ are the GT classification and the classification produced by the student, respectively, and $\lambda$ is a hyperparameter used to weight the loss terms. However, as previously stated, the direct usage of $p_t$ and $p_s$ to perform the KD step may not constitute a big difference from the usage of the GT labels, leading to small improvements in performance. In fact, since the softmax probabilities are extracted from a teacher with good performance, $p_t$ is expected to contain one value very close to one and the remaining values very close to zero \cite{duong2019shrinkteanet}. By considering this information directly, slight differences in perception between the non-selected classes will be diluted by the exponential function, eliminating domain knowledge that could be relevant for the student. This problem can be addressed by using a softer version of the logits that provides a better separation of the lower probabilities \cite{boutros2020compact,wu2020learning,boutros2022template}, which will be mentioned as \textbf{soft-response-based KD} (SRB-KD) throughout this document. The probabilities are softened by introducing a temperature parameter, $\tau$, in the softmax function (Equation \ref{eq:logits_noT}): \vspace{-1em}

\begin{equation}
    p_{m,i} = \frac{e^{\frac{z_i}{\tau}}}{\sum_{j=1}^{N}{e^{\frac{z_j}{\tau}}}}, m \in {t, s}, \tau > 1
    \label{eq:logits_T}.
\end{equation}

When FB-KD algorithms are used in supervised approaches, a classification loss function, $L_{class}(.)$, is usually considered in the model's loss computation since the KD term of the loss does not lead to its direct interpretation. Besides, two transformation functions, $\phi_t(.)$ and $\phi_s(.)$, should be applied when the teacher and student feature maps have different sizes, to ensure that comparable parameters are passed to $L_{KD}(.)$. Therefore, FB-KD loss can be written as:\vspace{-1.2em}

\begin{equation}
    L_{model} = L_{class}(c_{GT}, c_s) + \lambda L_{KD}(\phi_t(f_t), \phi_s(f_s)),
    \label{eq:fb_simple}
\end{equation}
where $f_t$ and $f_s$ are the features that the teacher and the student extracted from the layer involved in the distillation process, $\phi_t(f_t)$ and $\phi_s(f_t)$ have the same dimensions and $c_{GT}$, $c_s$ and $\lambda$ are defined as in Equation \ref{eq:rKD_class}. If the teacher's knowledge is transferred from $M$ layers instead of one, the previous equation can be rewritten as:\vspace{-1.2em}

\begin{equation}
    L_{model} = L_{class}(c_{GT}, c_s) + \sum_{i=1}^{M}\lambda_{i} L_{KD}(\phi_{t_i}(f_{t_i}),\phi_{s_i}(f_{s_i})),
\end{equation}
where $i$ represents the $i$-th pair of teacher-student layers entering the KD step.

In R-KD, two types of distillation can be performed with similar outcomes. On one side, the relation can be established between the teacher and the student and then compared with a specified target relation. This happens when the magnitude constraint present in FB-KD is dropped and only the direction of the teacher's and student's feature vectors are aligned \cite{duong2019shrinkteanet,boutros2022template,aslam2023privileged}. In this case, Equation \ref{eq:fb_simple} can be used for R-KD. On the other side, the relational information can be determined individually for the teacher and the student and, afterward, compared between them \cite{liu2022coupleface,caldeira2023unveiling}. This type of distillation requires that at least two feature vectors are considered for both architectures, as relational information is being extracted for each of them. However, it comes with the advantage of not comparing both architectures directly, eliminating the need for the transformation functions $\phi_t(.)$ and $\phi_s(.)$. In this case, Equation \ref{eq:fb_simple} can be rewritten as:\vspace{-1.2em}

\begin{equation}
    L_{model} = L_{class}(c_{GT}, c_s) + \lambda L_{KD}(f_t, a_t, f_s, a_s),
    \label{eq:realtion_KD}
\end{equation}
where $a_t$ and $a_s$ represent auxiliary teacher and student features with each $f_t$ and $f_s$ establish a relation, respectively. These auxiliary features are usually associated with each $f_t$ or $f_s$ in predefined pairs \cite{liu2022coupleface,caldeira2023unveiling}.

\subsection{Pruning} \label{pruning_intro}

Although in a much more intrinsic way, pruning techniques are also based on the human learning procedure. At the beginning of one's life, several neuronal connections are established to ensure a proper information flow between every part of the baby's body and brain \cite{chechik1998synaptic,zukerman2011brain}. In this scenario, the priority of the human body is to ensure that the flow occurs with minimal loss of information and, thus, several neurons are used to build the neuronal network, resulting in highly redundant connections that have a low contribution to the overall performance of the network. At the early stages of life, however, these connections are deemed to be necessary, since they guarantee a low amount of inconsistencies and flaws, which is of utmost importance for a child who is in contact for the first time with all the external sources of stimuli present in their surroundings. As the person grows up, however, processing external information becomes a standard task whose slight drop in performance will not majorly affect further evolution. Hence, later in life some of these neuronal connections are deactivated. This process, known as neuronal deactivation or synaptic pruning, is responsible for reducing the time needed to process external information and stimuli and the amount of memory and energy costs associated with the information flow without compromising it \cite{chechik1998synaptic,zukerman2011brain}.

\begin{figure}
    \centering
    \includegraphics[scale=0.48]{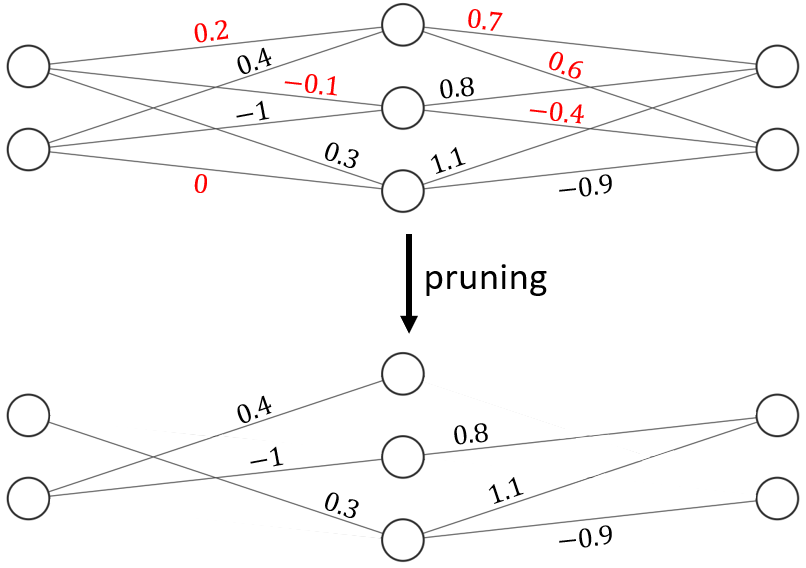}
    \caption{Pruning framework based on the traditional $L_1$-norm. The original network (top) is pruned in a layer-wise fashion to a sparsity of 50\%; the pruned weights are marked in red.}
    \vspace{-1.5em}
    \label{fig:prun}
\end{figure}

Pruning strategies work similarly. When a neural network is trained, it is important to have a high number of learnable parameters that ensure a high degree of freedom. Although some precaution is needed to avoid overfitting, the usage of overparametrized networks is often mandatory to ensure good model performance, since smaller networks are not able to properly learn the statistical distribution of the training data. After the model is trained, however, it is usually possible to reduce the total number of connections without highly impacting the final performance because some of them extract irrelevant or redundant information, as happens in the human body. Hence, pruning strategies compress networks by tearing apart some of their parts \cite{beyer2022knowledge}, resulting in sparse models as the one depicted in the bottom row of Figure~\ref{fig:prun}.

Four different factors need to be considered when performing pruning: the desired sparsity level, the granularity (LW, CW), the pruning criterion (L1-norm, for example) and the pruning strategy (SSPR, IPR). 

\subsubsection{Sparsity Level}
The percentage of connections to be removed defines how sparse the pruned network is. As such, a higher \textbf{sparsity level} represents more connections removed. The fact that the sparsity level can be decided by the user allows for a high degree of flexibility in balancing the size-performance trade-off, which constitutes a huge advantage when compared with less flexible techniques, such as quantization. While some applications might require a higher level of compression, others are less resource-constrained. With pruning, all types of networks can be derived from the same initial model by simply varying the sparsity level. Although sparser networks usually lead to a larger decrease in performance, some techniques have sparsity levels above 90\% with marginal performance drops or even improvement \cite{lee2018snip,lin2022fairgrape}. The reasons leading to this performance improvement are discussed in detail in Section \ref{sec:pruning_SOTA}.

\subsubsection{Pruning Granularity}
Pruning can be applied at different granularities, as in quantization, as long as a continuous flux of information is maintained. Since eliminating a whole layer would break the information flow, group-wise pruning is not used. However, as in quantization \cite{bunda2022sub}, it is possible to prune each layer/channel to a distinct sparsity level \cite{li2016pruning}. Hence, pruning strategies can be divided as follows:

\begin{itemize}
    \item \textbf{layer-wise (LW) pruning}: pruning is individually applied to the channels/weights within each layer. When the selected pruning percentage does not lead to a final integer number of channels/weights, the rounding strategy should ensure that at least one is preserved in every situation \cite{polyak2015channel,liu2021discrimination,li2016pruning,zhu2017prune}.
    \item \textbf{channel-wise (CW) pruning}: pruning is individually applied to the kernels within each channel, considering the rounding constraint mentioned above \cite{liu2021discrimination,vitek2023ipad}.
\end{itemize}

\subsubsection{Pruning Criterion}
Although the weights to be pruned within each pruning group could be randomly selected, it is essential to choose a pruning criterion that does not compromise overall performance. As such, each group of weights is usually ordered in a predefined way and the weights with lower scores are eliminated until the desired sparsity level is reached. The most common ordering criterion is the $L_1$-norm since near-zero-valued weights usually do not contribute very significantly to the final output values. This measure can also properly represent the magnitude of the activations that follow a certain set of connections without the need to use a calibration set \cite{li2016pruning}. Making the pruning process data-independent removes the uncertainty associated with the selection of an appropriate calibration set, resulting in a deterministic pruning criterion. Furthermore, this strategy leads to lower performance drops than random pruning \cite{li2016pruning,liu2021discrimination,vitek2023ipad,yu2018nisp}, proving that selecting an appropriate pruning criterion is of the utmost importance. Hence, the study of new and efficient pruning criteria is essential to ensure that pruning techniques are exploited to their full potential. 

\subsubsection{Pruning Strategy}
As happened in QAT, pruning can be followed by retraining so that the network adapts to its new architecture and regains some of the lost performance. A common approach is to simultaneously prune each layer or channel to the desired sparsity level and, once the whole network has been pruned, retrain until convergence \cite{li2016pruning,lee2018snip}. It is also possible to prune each layer or channel in different moments separated by a retraining phase \cite{polyak2015channel}. This way, the model is retrained after each predefined group suffers pruning, adapting to the already imposed changes before any further pruning is applied. Although \textbf{iterative pruning and retraining} (IPR) strategies usually achieve better results, this type of procedure is extremely time-consuming when compared with \textbf{single shot pruning and retraining} (SSPR) due to the existence of several retraining steps, especially when very deep networks are being considered \cite{li2016pruning}.

As previously stated, pruning is a flexible way of compressing a model since it allows the user to select the desired sparsity level according to the computational restrictions of the task at hand. However, this compression strategy presents some challenges that should not be disregarded, namely the sparsity of the resultant networks. When a compression strategy like quantization is applied, the compressed network remains dense, which allows computations to be executed in the exact same way as in the original model. When pruning strategies are applied some connections within the network are eliminated and the model becomes sparse, which induces irregularities in the data structures that are used to perform computations at inference time. The existence of these irregularities is not usually compatible with standard hardware and requires the usage of specific hardware to be deployed, which poses a disadvantage when compared to models resulting from other compression techniques, that can be easily and efficiently deployed in different types of hardware \cite{jeon2022mr}. Besides, pruning a network changes its architecture in a way that cannot be predicted beforehand, meaning that its architecture-dependent components (such as batch normalization layers) might be affected by the pruning process \cite{beyer2022knowledge}.

\section{State of the Art}\label{sec:sota}
\subsection{Quantization}

This section describes the current literature on quantization techniques in the fields of biometrics and CV. A summary of the analyzed studies can be found in Table~\ref{tab:quantization}.

\subsubsection{Quantization in Biometrics}

Quantization can be performed in models designed for several tasks, namely in biometrics. Boutros~\textit{et al.}~\cite{boutros2022quantface} proposed a novel solution that integrated model quantization and KD to compress FR models up to five times using synthetic data. The authors followed a WAQ strategy that allowed them to compare the performance of the original model to its 8 and 6-bit quantized versions. To minimize performance drops associated with the quantization, the experiments followed a QAT framework where the quantized model (student) was fine-tuned through KD from the FP model (teacher) at the images' feature embeddings level (FB-KD). The value of $L_{KD}$ was computed considering the cosine distance as a similarity measure between the normalized embeddings. Although the quantization process resulted in small performance drops, the high memory gain makes these losses acceptable to some extent, proving that it is possible to regulate the computational cost of the FR model without severe performance degradation.

Bunda~\textit{et al.}~\cite{bunda2022sub} addressed the problem of sub-byte quantization ($b<8$) in a WAQ FR framework by comparing the results obtained using this method on a modified MobileFaceNet \cite{chen2018mobilefacenets} with the ones achieved by its FP and byte-quantized ($b=8$) versions. Although 2-bit quantization performed poorly, 4-bit quantization achieved a good performance after fine-tuning with QAT, proving that sub-byte quantization does not necessarily lead to high performance drops. This study also analyzed mixed-precision quantization (MPQ), a procedure where different sets of parameters are quantized with a different number of bits. Using a lower number of bits to represent the weights and activations of the layers that do not have a big negative impact on the performance can highly reduce the computational resources needed at inference time without significant performance degradation. The authors prove that model performance suffers a bigger impact when the first layers are quantized due to the forward propagation of the performed approximations, allowing for stronger quantization when progressing toward the last layers of the network. The final MPQ framework comprised 8, 4 and 2-bit quantization for the initial, middle and final layers, respectively, resulting in a network with only 10\% of the computational footprint of the original model and a performance drop of 0.68 percentual points on LFW. Thus, MPQ can be extremely useful in establishing a compromise between achieving good performances and having computationally efficient models. However, determining the MPQ configuration manually will most likely lead to a suboptimal quantization, especially considering the exponential growth of possible configurations with the network size \cite{gholami2022survey}, revealing the need to develop methods that automatize the MPQ configuration selection during quantization. 

Kolf~\textit{et al.}~\cite{kolf2022lightweight} focused on the compression of models trained on the periocular recognition (PR) task through a QAT procedure that performed WAQ at 8, 6 and 4-bit precision. The achieved results show that the more the number of bits of the final representation is reduced, the bigger the verified performance drop, as expected. When $b=8$ or $b=6$, the performance degradation was small in the identification and verification tasks and, in some cases, the quantized models even slightly surpassed the baseline. For $b=4$, the performance suffered an extreme drop. As an example, a MobileFaceNet trained in the identification task suffered a reduction in accuracy from 99.80\% in 6-bit quantization to 5.53\% in 4-bit quantization. Hence, in this experimental setup, 4-bit quantization is not enough to achieve good performance, showing that this precision level does not allow the model to learn how to encode the relevant features properly. 

Kolf~\textit{et al.}~\cite{kolf2023syper} also compressed PR models in a WAQ framework at 8, 6 and 4-bit precision. The FP models were trained on visible spectrum images, which are widely available and labeled. To analyze the impact of using distinct datasets to perform QAT, the authors compared models quantized using the training data of the FP model, a GAN-generated synthetic dataset of visible spectrum images, SyPer, and a dataset with real near-infrared (NIR) images. In the two last scenarios, QAT was performed through FB-KD to assess whether quantization with privacy-friendly information impacts the quantized model's performance. QAT based on SyPer resulted in reduced performance drops, especially for 8 and 6-bit quantization. It was comparable to the drop resultant from QAT with the original training set, proving that privacy-friendly data can be used in quantization frameworks without enlarging the expected performance degradation. Models quantized with NIR data surpassed the ones obtained using SyPer and outperformed their FP versions for $b>4$, highlighting the possibility of using quantization as a way of adapting pre-existing models to new input domains. This is particularly important when the new domain has a lower amount of data available.

In an additional effort towards improved quantization, Kolf~\textit{et al.}~\cite{IJCB2023JanColor} has also analyzed the impact of combining color quantization (CQ) and model quantization in FR systems performance. CQ reduces the space occupied by the FR model's reference data by reducing the number of bits that represent each color channel ($b_{CQ}=8$ without CQ). CQ alone resulted in reduced performance drops when $b_{CQ}=4$ and $b_{CQ}=3$; for $b_{CQ}=2$, however, model performance dropped significantly. To analyze the impact of the data used to perform QAT, the CQ-based models were quantized using both not-quantized and CQ data. The usage of CQ data resulted in improved model performance in comparison with non-quantized data, especially $b_{CQ}=3$ or $b_{CQ}=2$. Hence, the data used for QAT should be carefully selected, as it can highly impact the quantized model's performance \cite{kolf2023syper}. Furthermore, the quantized models surpassed their FP versions when dealing with CQ input data, especially when $b_{CQ}=2$. Quantized models are inherently more capable of dealing with quantized input data since they work with lower-bit information by default, allowing for a reduction of 75\% and 81\% of the required memory for face images' storage and of the model size, respectively, without severe performance degradation. 

\subsubsection{Quantization in Computer Vision} 

Quantization techniques have been studied in greater depth in computer vision tasks than in the biometrics field. Zhou~\textit{et al.}~\cite{zhou2017incremental} proposed a novel WOQ strategy that avoids quantizing all the weights of the network in a single step to disturb its performance as little as possible at each quantization step, resulting in an iterative PTQ (iPTQ) methodology. The authors divide the training strategy into three steps that are iteratively repeated until complete weight quantization is achieved. First, the FP weights of each layer are divided into two disjoint groups with predefined sizes. Then, the weights in the first group are quantized in a LW fashion, while the weights in the second group remain unchanged. Finally, the model is re-trained by updating the set of FP weights to compensate for the quantization-induced accuracy drop; the already quantized weights remain fixed. Their 5-bit quantized network achieved better results than the evaluated SOTA references without requiring a high number of re-training epochs. To achieve further compression, this strategy could be deployed to a WAQ framework, as proposed by the authors. 

Jacob~\textit{et al.}~\cite{jacob2018quantization} focused on enhancing the trade-off between latency and accuracy while addressing two problems. On one side, a large number of works use over-parameterized networks as the FP base of the quantized model. To step up to more challenging scenarios, Jacob~\textit{et al.}~\cite{jacob2018quantization} used MobileNets \cite{howard2017mobilenets}, which are small but efficient in the latency versus accuracy trade-off. On the other side, they also verified how effective their quantization strategies are when deployed in real hardware, which is often lacking in previously published studies (such as Wang~\textit{et al.}~\cite{wang2022learnable} and Choi~\textit{et al.}~\cite{choi2020data}). Since PTQ usually results in big performance drops when applied to small models \cite{bunda2022sub,jacob2018quantization}, a QAT scheme was selected to perform WAQ to 8-bit precision. On the image classification (IC) task, the quantized MobileNets outperformed their FP versions when given a fixed run-time budget. This type of experiment is particularly useful to evaluate the quality of the produced previsions on real-time applications. Despite showing that 8-bit precision models can be about 10 percentual points better than its FP version in these circumstances, the authors also state that the observed difference depends on the selected hardware, being smaller when using hardware that can better handle FP computations. 


Although predefined quantization functions have been widely used to compress FP models, they might not constitute the best way to minimize the quantization error. The optimal quantizer should lead to the lowest quantization error achievable with the considered input data distribution, which most likely does not follow a function that can be determined beforehand. Zhang~\textit{et al.}~\cite{zhang2018lq} proposed learnable WAQ and WOQ strategies that can be applied to any architecture. The quantization operations are designed as inner products between a binary and a coding vector whose entries are learned to adapt to each layer (activations) or channel (weights) of the model during training. WOQ led to very small performance drops when compared with the original FP models, even for 3-bit precision, while WAQ led to more significant performance drops in general. When compared with the literature, this method resulted in improved performance both in WOQ and WAQ. Furthermore, the authors show that the learned quantizers are not uniform and vary between layers/channels, proving that uniform quantizers are often sub-optimal. Thus, learnable quantization techniques should be further investigated for deployment in resource-constrained environments.

Jeon~\textit{et al.}~\cite{jeon2022mr} proposed a learnable PTQ framework that quantizes weights (CW) and activations (LW) simultaneously, instead of sequentially. Differently from what happened in Zhang~\textit{et al.}~\cite{zhang2018lq}, this framework does not minimize the difference between the scaled versions of FP and quantized weights (quantization error) but the difference between the output values of each quantized set of parameters for the FP and quantized models. The reasoning behind this choice is that using an optimization procedure that considers the input data distribution might be beneficial to the model's performance. It is interesting to note that using the reconstruction error to optimize the model can be considered a KD process where the FP model acts as the teacher of the quantized student. The experiments performed with CNNs, Transformers \cite{vaswani2017attention} and ViT \cite{dosovitskiy2020image} networks showed that this method overcomes the existent SOTA references in every situation. The benefits associated with this methodology are particularly high when low-bit precision quantization is being performed, despite that most of the developed LP models are still very far away from FP accuracy. Hence, this study highlights the importance of expanding research in the field of learnable quantizers.

Wang~\textit{et al.}~\cite{wang2022learnable} presented a learnable quantization technique based on cost-effective lookup tables that learn how to map each FP parameter in the quantization spectrum. These lookup tables are interpreted as graphs where the x-axis represents the FP spectrum. The y-axis is uniformly divided into $2^b$ areas, representing the quantization spectrum. The final mapping can then be represented by a set of $2^b-1$ impulses with the same magnitude whose placement in the x-axis is learned during the training procedure. Although this placement will most likely be non-uniform, the model forces values that can be quantized without error to be quantized that way, reducing the problem to the placement of a single impulse in each of the $2^b-1$ subdivisions of the y-axis. Models resultant from layer-wise WOQ and WAQ were tested in IC, super-resolution (SR) image generation and point cloud classification (PCC), consistently maintaining comparable accuracy to their FP versions and outperforming SOTA references. Moreover, the verified reductions in activation quantization time (57.8\% in mobile processors and 76.5\% in CPUs) prove the usefulness of learnable lookup tables in resource-constrained settings.

Even though several studies present models that achieve good performances after compression, one should be careful when analyzing their results. For instance, Ji~\textit{et al.}~\cite{ji2022neural} proposed a compression approach using KD and quantization for a bearing fault diagnosis (BFD) model. An SRB-KD strategy is used to distill knowledge from a teacher to a much simpler student, which is then quantized for further compression. Despite being an interesting framework, the performed analysis might be misleading if not properly examined. When conducting a study to assess the influence of the KD step on the student's performance, the authors conclude that this strategy is responsible for a remarkable performance rise of 8.24 percentual points. However, this comparison is established with a non-distilled student that was not trained for the same number of epochs, which can lead to an overestimation of the positive effects of the KD step. This shows that the careful analysis of both the results and the conditions that allowed to obtain them should never be disregarded.

\begin{table*}[]
    \caption{Summary of the literature on quantization. It can be noted that all the analyzed biometrics works used QAT. \textbf{Task}: FD: face detection, OD: object detection, NLP: natural language processing, VT: vision tasks; *CW and LW quantization applied to the weights and activations, respectively}
    \centering
    \begin{tabular}{c|c|c|c|c|c|c}
         \textbf{Area} & \textbf{Paper} & \textbf{Year} & \textbf{Task} & \textbf{Granularity} & \textbf{Strategy} & \textbf{$b$} \\
         \hline
         \multirow{5}{*}{Biometrics} & \cite{boutros2022quantface} & 2022 & FR & CW & WAQ, QAT & 8, 6\\
         & \cite{bunda2022sub} & 2022 & FR & LW & WAQ, PTQ/QAT & 8, 4, 2\\
         & \cite{kolf2022lightweight} & 2022 & PR & CW & WAQ, QAT & 8, 6, 4\\
         & \cite{kolf2023syper} & 2023 & PR & CW & WAQ, QAT & 8, 6, 4 \\
         & \cite{IJCB2023JanColor} & 2023 & FR & CW & WOQ, QAT & 8, 6\\
         \hline \hline
         \multirow{6}{*}{CV} & \cite{zhou2017incremental} & 2017 & IC & LW & WOQ, iPTQ & 5, 4, 3, 2 \\
         & \cite{jacob2018quantization} & 2018 & IC, OD, FD, FAC & LW & WAQ, QAT & 8\\
         & \cite{zhang2018lq} & 2018 & IC & CW, LW* & WAQ/WOQ, PTQ & 4, 3, 2, 1 \\
         & \cite{jeon2022mr} & 2022 & IC, VT, NLP & LW, CW & WAQ/WOQ, PTQ & 8, 6, 4, 3, 2 \\
         & \cite{wang2022learnable} & 2022 & IC, SR, PCC & LW & WAQ/WOQ, PTQ & 8, 4, 3, 2\\
         & \cite{ji2022neural} & 2022 & BFD & LW & WOQ, PTQ & 8  
    \end{tabular}
    \label{tab:quantization}
    \vspace{-1.5em}
\end{table*}
\subsection{Knowledge Distillation} \label{sec:KD_SOTA}

This section describes the current literature on KD techniques in the field of biometrics. A summary of the analyzed studies can be found in Table~\ref{tab:KD}.

\subsubsection{KD in Biometrics}

KD algorithms have already been used to aid the development of lighter models in the biometrics area, namely in FR. Boutros~\textit{et al.}~\cite{boutros2022pocketnet} used a multistep KD (MS-KD) methodology to distill knowledge to a lightweight student network used in FR applications with limited memory specifications. MS-KD consists of training the teacher and the student models side by side so that the student can keep up with the teacher's learning process. Instead of contacting with the embeddings produced by an already trained teacher, the student gets access to the embeddings produced by the teacher during its own training phase, and the knowledge is periodically transferred during their simultaneous training. The proposed algorithm defines $L_{KD}$ as the mean squared error (MSE) loss and compares the MS-KD performance with the ones obtained without KD and with classic KD, proving that MS-KD enhances FR performance.

Wang~\textit{et al.}~\cite{wang2021teacher} studied FB-KD, defining $L_{KD}$ as the $L_2$ loss between the teacher and student's features. The validation loss comprises an extra term that estimates the computational cost of the defined model, $L_{comp}$. This allows for the utilization of Neural Architecture Search (NAS) \cite{zoph2016neural} to optimize the student's architecture. The utilization of NAS coupled with KD can be extremely useful since it helps adapt the student to the knowledge that is being transferred, taking better advantage of it. Another advantage of this type of definition is that it allows the user to predefine the desired complexity of the student, measured by $L_{comp}$, adapting the resultant network to the level of compactness required by its application scenario. Hence, Wang~\textit{et al.}~\cite{wang2021teacher} simultaneously trained the student to approximate the teacher's feature space and to adapt its architecture to improve the optimization procedure while approximating the desired compression level. The obtained results are in line with the expectations, as using KD proved to be better than training the compressed model from scratch and FB-KD usually outperformed RB-KD.

Luo~\textit{et al.}~\cite{luo2016face} also used a simple $L_2$ loss between the features produced by a teacher and a student in the FR task. However, this loss term does not comprise all the neurons of each layer, since the neurons in the middle of the network encode information regarding different characteristics of the analyzed faces. This information includes identity-related attributes such as race and gender, which are relevant for the FR task, and non-related attributes such as illumination, that should not influence the FR algorithm \cite{luo2016face}. Hence, the selection strategy penalizes neurons that contain irrelevant information (mainly non-related attributes) and pairs of neurons that encode similar information (since the same identity-related attribute can be encoded by more than one neuron, which is redundant). This strategy produced faster and simpler models without severe performance degradation. In some cases, the student was even able to surpass the teacher, proving that neuron selection can preserve important information while reducing the influence of the noise learnt by the teacher. By using teachers with different complexity levels, the authors were also able to conclude that the teacher-student gap plays a major role in KD efficiency.

Li~\textit{et al.}~\cite{li2023rethinking} addressed the teacher-student gap problem by studying its correlation with each model's intrinsic dimension, that is, the minimum number of variables it needs to describe all points in its feature space without any ambiguities. Intrinsic dimension can also be used to describe the model's ability to generalize against irrelevant information for the task at hand, such as noise. The lower the intrinsic dimension, the less noise can be found in the feature space, making its unambiguous description easier. As models with more parameters are usually associated with lower intrinsic dimensions, the authors used a reverse distillation procedure that transferred knowledge from the student to the teacher to impose some constraints in its feature space, increasing its intrinsic dimension and, thus, closing the intrinsic gap. Since the feature space of the teacher is constrained by the student’s one, it is easier for the final student to mimic it. Hence, the first student is trained in a labeled dataset with an Arcface loss function. The teacher is then trained on the same dataset, considering the classification loss and an extra term that performs reverse distillation through the features extracted by the student. After training, the teacher model pseudo labels an unlabeled dataset. The extracted features are used to perform FB-KD to a new student model. When using the same architecture for the initial and final students, the proposed method was extremely effective. The obtained teachers presented a higher intrinsic dimension and, thus, lower intrinsic gap regarding the student. All the analyzed student networks suffered considerable performance improvements and converged faster when this methodology was used, suggesting that it was easier for them to mimic the teacher's behavior, as expected. 

Huang~\textit{et al.}~\cite{huang2022evaluation} escaped the rigidness of the classic KD methodologies by applying a novel R-KD method that does not require any correlation between teacher and student architectures. It is done by defining a set of thresholds that separate the pairs of mismatched identities within the training dataset according to whether they are misclassified or not when a certain False Positive Rate (FPR) is pre-defined. These thresholds are defined separately for the teacher and the student and allow to define a set of critical pairs, that is, pairs that fall between thresholds associated with different FPRs for both networks, contributing to the disparity between their results. Since critical pairs are the only ones contributing to the loss computation, there is a relaxation of the approximation of the architectures, removing any restriction on the student's architecture. Hence, the authors were able to compare two models with different complexities: a ResNet-18 \cite{he2016deep} and a MobileFaceNet \cite{chen2018mobilefacenets}. The verified performance improvements were greater for the student with lower capacity (MobileFaceNet), which aligns with the goal of developing a strategy that can be used to distill knowledge to compressed models.

Boutros~\textit{et al.}~\cite{boutros2020compact} proposed a KD approach to enhance the performance of mobile phone periocular image processing with compressed models. $L_{KD}$ was defined as the KL divergence between the soft version of the logits of the teacher and the student (SRB-KD). The authors compared four models with the same architecture trained following different strategies: direct training without KD and training with knowledge distilled from three teachers of different complexities. Besides showing that the KD models surpass the student trained without KD, this study shows that the KD student with better overall performance is the one whose training was guided by the teacher architecture with a lower amount of parameters. This may be because the soft logits produced by the more complex teacher consider complex information that the student is not able to interpret, increasing its difficulty to converge, which agrees with the conclusions withdrawn by Luo~\textit{et al.}~\cite{luo2016face} and Li~\textit{et al.}~\cite{li2019graph}.
 


Boutros~\textit{et al.}~\cite{boutros2022template} further addressed PR with compressed models. The authors aimed to prove that using both RB-KD and FB-KD or R-KD can consistently outperform the usage of RB-KD alone. They compared the classification performance of an RB-KD method conceptually similar to the one presented by Boutros~\textit{et al.}~\cite{boutros2020compact} with two models resultant from adding a loss term to $L_{KD}$, one based on MSE (FB-KD) and another based on CS (R-KD). The models that used the two types of KD outperform the RB-KD architecture, showing that the embeddings produced in the inner layers contain useful information that cannot be extracted from the logits, even when their soft versions are being used. Neither MSE nor CS produced consistently better results than the other. Although this might seem inconclusive, the fact that CS is competitive or better than MSE shows that it is possible to relax the impositions made on the KD loss terms without performance degradation. While MSE tries to force an exact match between the features extracted by the teacher and the student, CS drops the magnitude constraint. This allows for higher freedom of the student output, making it easier to adapt its feature space to resemble the teacher's.

Duong~\textit{et al.}~\cite{duong2019shrinkteanet} followed a similar line of thinking, stating that forcing the student and teacher embeddings to exactly match will most likely lead to an unstable and over-regularized model. The authors proposed the alignment of the embeddings produced by both models in the same direction, disregarding their magnitude, as happened in the model based on CS of Boutros~\textit{et al.}~\cite{boutros2022template}. The developed strategy allows the selection of the layers used for embedding comparison; when a layer different from the last one is picked, the student embedding is converted into the teacher space and passed through its frozen architecture to produce the final representation, which is compared with one retrieved by the teacher. In other words, this strategy reflects the ability of the teacher model to retain its predictions stable if it relied on the embedding produced by the student instead of its own. Hence, Duong~\textit{et al.}~\cite{duong2019shrinkteanet} proposes an interesting and creative method, different from the general strategies followed in KD architectures.

Wu~\textit{et al.}~\cite{wu2020learning} focused on model acceleration rather than compression. The utilized $L_{KD}$ comprised a SRB-KD and a FB-KD loss terms. The SRB-KD term utilized the $L_1$ loss while the FB-KD term between each student sample and the center of its class resulted in an angular loss term. The idea behind this strategy is that open-set problems like FR require the student to mimic the embedding space of the teacher, rather than the produced labels, as the identities present in the train and test sets are most likely different from the ones that the model will need to identify in real-world applications. The proposed loss and architecture highly accelerated the student models while maintaining good performance in several open-set tasks. The students often surpassed SOTA performances and the teachers used to train them, proving that KD should not be limited to model acceleration or compression. Besides, the obtained results corroborate the conclusions of Boutros~\textit{et al.}~\cite{boutros2022template} and Duong~\textit{et al.}~\cite{duong2019shrinkteanet}, by proving that it is important to explore KD terms with different levels of flexibility since using extremely constrained KD losses can lead to poor results. 

The traditional $L_2$ loss usually considered in FB-KD applications can be misleading as it approximates the teacher and student's embedding spaces without guaranteeing that the distance between samples is preserved. Preserving this distance is of extreme importance in FR tasks, since the separation between pairs of samples allows the classifier to distinguish between identities. This can both come as an advantage (if the student's features of an impostor pair are more separated, for example) or a disadvantage (in the opposite case). Taking this into consideration, Liu~\textit{et al.}~\cite{liu2022coupleface} proposed an R-KD method based on a CS loss that does not disturb the impostor pairs where the student has an advantage over the teacher while incorporating the remaining ones in the loss, as they can benefit from the teacher's knowledge. This extra $L_{KD}$ term highlights the impact of hard negative samples for improved discriminative power. This method surpassed the traditional feature-based $L_2$ loss, proving that focusing the optimization process on the disparities that are prejudicial to the student's performance can lead to an enhancement of KD-derived models. This approach is theoretically simple, posing an elegant solution to some of the problems faced by KD.

Zhao~\textit{et al.}~\cite{zhao2023grouped} also presented a KD strategy designed to transfer the most relevant and meaningful knowledge to the student. This RB-KD approach divided $L_{KD}$ into three independent terms: primary, secondary and binary losses. The primary loss is extracted from the output neurons that retrieve higher probability values for a specific sample, as they are more likely to contain relevant information. The secondary term encompasses the remaining neurons. The binary loss term preserves the consistency of the knowledge distribution between the teacher and the student. As shown in the experiments both primary and binary terms are essential for a strong student performance. Contrarily, the secondary term negatively impacted the performance of the student. This may happen because the information encoded by the secondary term is mainly useless for the training procedure. Relaxing the problem by removing the influence of the small valued output neurons may allow the student to better focus on the remaining information. This method surpassed several SOTA models, namely Huang~\textit{et al.}~\cite{huang2022evaluation}, which follows a FB-KD approach. The proposed strategy can also be generalized for other KD application scenarios, namely masked FR, despite not being able to surpass Huber~\textit{et al.}~\cite{huber2021mask}. Combining the losses proposed in these two papers could present an interesting research direction since RB-KD can help improve the performance of models trained with FB-KD \cite{ge2020efficient}. A possible way of deploying this study is to adapt the presented idea to FB-KD frameworks. Although the separation between the primary and secondary terms is not as trivial in this scenario, it could be interesting to decide which features should be distilled or ignored for each sample, using, for instance, saliency maps similar to the ones usually employed in xAI applications \cite{neto2022explainable}.  

Xu~\textit{et al.}~\cite{xu2023probabilistic} used KD to generate a student model from an ensemble of teachers, greatly reducing the memory and computation costs of generating predictions. While using a single teacher might lead to undesired situations (e.g. the distillation of biased predictions), their effects can be averaged by distilling knowledge from several sources. Furthermore, the ensemble members do not have the same ease in classifying each sample, meaning that its perceived quality is different according to the model used to classify it. This supports the usage of a weighted average between the features produced by the ensemble members instead of a simple arithmetic average. Hence, the teacher networks are addressed from a probabilistic point of view, resulting in a method titled Bayesian Ensemble Averaging that transfers feature-based knowledge while measuring the samples' quality. Using it to perform KD resulted in better performance than using a simple average that disregards the differences between each model of the ensemble.

As previously stated, the usage of KD algorithms is not restricted to model compression and acceleration tasks. Huber~\textit{et al.}~\cite{huber2021mask} aimed to bridge the accuracy gap of FR models in masked faces, proposing a mask-invariant FR technique based on KD. Two models with the same architecture were trained on the recognition task. They were fed the same samples in the same order but the student model's input images were covered with a synthetic mask half of the times. This way, the model was trained to classify both masked and unmasked faces. The KD step compared the embeddings produced by both models since images from the same identity should produce embeddings as similar as possible, whether the person uses a mask or not. Hence, the usage of KD ensures that the extracted features are minimally influenced by the presence of a mask. The proposed algorithm defined $L_{KD}$ as the MSE loss and two KD approaches were tested: one that used a fixed value of $\lambda$ and another that increased it after the stabilization of $L_{class}$, to further remove the model's attention from the mask. The latter approach resulted in a big decrease in the value of $L_{KD}$, proving its efficacy. Besides, KD was beneficial to the verification problem, improving masked FR without a significant drop in performance for unmasked faces. 

Unsurprisingly, the importance of the problem addressed by Huber~\textit{et al.}~\cite{huber2021mask} was highlighted after the beginning of the COVID-19 pandemic. However, FR systems have suffered from other problems regarding image acquisition conditions. FR algorithms are usually trained and tested using high-resolution images from publicly available datasets. However, the images that these models are designed to classify are commonly acquired in poorer conditions (for instance, long-distance images acquired by surveillance cameras), resulting in significant performance drops that make the model inappropriate for real-world scenarios. One way to mitigate this is by training models specialized in recognizing low-resolution images. Using this type of image to directly train the final model generally results in reduced recognition performance since the relevant identity information encoded in low-resolution images is hard to access without proper guidance. This problem can be addressed with KD by training the teacher in high-resolution images, which makes it able to extract the more relevant traits for the FR \cite{ge2018low,ge2020efficient,boutros2022low}. Then, the student is trained in low-resolution images with the teacher's supervision, ensuring the desired domain adaption. If the student's architecture is simpler than the teacher's, this methodology can be simultaneously used to achieve compression.

Ge~\textit{et al.}~\cite{ge2018low} used this idea to develop a low-resolution FR system whose complexity is about 175 times inferior to its teacher's, a VGGFace \cite{parkhi2015deep}. The low-resolution images used to train and test the student result from resolution degradation of the high-resolution images that are fed to the teacher, followed by data augmentation. This way, the teacher can directly distill its knowledge to the student architecture. The authors utilize FB-KD, defining $L_{KD}$ as the $L_2$ loss between the embeddings extracted by both models. As this study highlights, the focus of KD techniques should be transferring accurate information between the studied networks since there is no use in distilling information that the teacher perceived incorrectly. Hence, $L_{KD}$ only considers the samples that are aligned (distance-wise) with the centroids of their class, thus removing the outliers' effect on the final model and prioritizing correct teacher predictions. These models proved to be competitive with SOTA architectures, while highly reducing the computational cost of the designed FR system. Besides, KD techniques surpassed the conventional methodologies in the low-resolution face retrieval task. In particular, in the challenging dataset SCface, all the tested KD methodologies surpassed the baseline, with impressive accuracy improvements (16.5\% with the baseline vs 48.33\% with their KD technique).  
 
Ge~\textit{et al.}~\cite{ge2020efficient} also addressed FR under low-resolution constraints. The process of obtaining low-resolution images was similar to the one used in Ge~\textit{et al.}~\cite{ge2018low} but they employed both RB-KD and FB-KD, with different goals. The RB-KD step adapts the knowledge learned by the teacher between two high-resolution datasets: the private one, which was used to train it, and a public one, which will then be used to transfer the acquired knowledge to the student. This step aims to adapt the teacher's knowledge to a domain that is closer to the student's while reducing the complexity of the produced feature space through compression. The adapted teacher is then used to distill its knowledge to a less complex student, with an $L_2$ FB-KD loss. This strategy was able to further reduce the complexity of the model presented by Ge~\textit{et al.}~\cite{ge2018low}. The RB-KD step reduced the teacher's memory consumption and increased the inference speed while enhancing the performance.

The usage of KD to enhance the performance of biometric models when dealing with low-resolution images is not restricted to FR tasks. Boutros~\textit{et al.}~\cite{boutros2022low} trained a model to perform iris recognition (IR) in low-resolution images. The teacher was trained using high-resolution iris images and its knowledge was distilled to the low-resolution trained student using FB-KD, in an approach that highly resembles the one followed by Huber~\textit{et al.}~\cite{huber2021mask}. As demonstrated in this study, Boutros~\textit{et al.}~\cite{boutros2022low} proved that the usage of KD can highly benefit the learning process of the student model.

Aslam~\textit{et al.}~\cite{aslam2023privileged} presented a KD methodology from multi-modal emotion recognition (ER) to single-mode ER. Multi-modal ER generally results in enhanced performance by considering several measures that can present complementary information. This is particularly relevant in situations affected by occlusion, where unoccluded channels may compensate for the occluded ones. However, the usage of multi-modal approaches is not always feasible, since some of the required data, such as physiological signals, might be unavailable upon inference time \cite{aslam2023privileged}. Hence, a multi-modal teacher that has both audio and visual inputs was used to transfer knowledge to a single-mode student that only processes visual information. Although this makes the student much lighter than the teacher, this problem should not be interpreted as a simple compression task since the main goal is to transfer knowledge from a multi-modal scenario to a more constrained single-modal framework. In a way, the fact that the student has access to limited information when compared with the teacher makes this problem resemble the one addressed by Huber~\textit{et al.}~\cite{huber2021mask}, for which KD proved to be advantageous. The KD procedure is performed by using CS to approximate the relation between the features produced by both models. This way, the student can learn to extract relevant information that is enhanced by the audio analysis without ever receiving audio as input. When the teacher’s prediction is not statistically similar to the GT, the KD loss term is disregarded by a negative transfer module. In the remaining cases, the KD term is given more importance when there is a high correlation between the teacher’s prediction and the GT, resulting in a KD with variable impact for each analyzed sample. The comparison between the obtained student and a baseline trained from scratch was made using two different backbones. In the less challenging experimental setting, using KD resulted in considerable performance improvement, especially when the negative transfer module was used, proving its usefulness. On more challenging conditions, however, the student was only able to perform well in one of the two extracted metrics, showing once more that the teacher-student gap should be taken into consideration before deploying a KD method.

Another interesting and distinctive proof of the usefulness of KD methodologies can be found in Caldeira~\textit{et al.}~\cite{caldeira2023unveiling}, a study focused on morphing attack detection (MAD) in FR. Based on the fact that \textit{bonafide} and morphed samples can be distinguished by the number of identities they encode, the authors proposed a methodology that distills identity information extracted by an autoencoder to the morphing classifier. The autoencoder is trained on the \textit{bonafide} samples that were fused to generate the morphing attacks. The morphing classifier extracts two vectors that contain information regarding the present identities and are, thus, used to predict two identity scores. For \textit{bonafide} images, the KD loss term approximates the identity information of the vector that is more aligned with the one produced by the autoencoder to 1, while approximating both the other identity score and its vector to 0. For attack samples, both vectors are expected to encode identity information and, thus, R-KD is used to distill the angle between them from the autoencoder through CS. This methodology surpassed other MAD architectures in the majority of the tested datasets. It also improved the interpretability of the morphing attack detection problem, as the analysis of the identity scores produced by each sample might help to explain its classification. By presenting a differentiated analysis of KD capabilities, this study proved the usefulness and importance of performing extensive research on this type of technique.

\begin{table}[]
    \caption{Summary of the literature on KD in biometrics. It can be noted that the amount of KD works available on biometrics is significantly higher than in quantization and pruning. \textbf{Type of KD}: RFB: reverse feature-based; \textbf{KD Loss}: AL: angular loss, CEL: cross-entropy loss, EKDL: evaluation KD loss, KLD: Kullback–Leibler divergence, N-$L_2$: normalized $L_2$, W$L_2$-MC: weighted $L_2$ for mean and concentration}
    \centering
    \begin{tabular}{c|c|c|c|c}
        \textbf{Paper} & \textbf{Year} & \textbf{Task} & \textbf{Type of KD} & \textbf{KD Loss} \\
         \hline
         \cite{luo2016face} & 2016 & FR & FB & $L_2$ \\
         \cite{ge2018low} & 2018 & FR & FB & $L_2$\\
         \cite{duong2019shrinkteanet} & 2019 & FR & R & CS\\
         \cite{boutros2020compact} & 2020 & PR & SRB & KLD\\
         \cite{wu2020learning} & 2020 & FR & SRB, FB & $L_1$ + AL\\
         \cite{ge2020efficient} & 2020 & FR & RB, FB & CEL, $L_2$\\
         \cite{wang2021teacher} & 2021 & FR & FB & $L_2$\\ 
         \cite{huber2021mask} & 2021 & FR & FB & MSE\\
         \cite{boutros2022pocketnet} & 2022 & FR & MS & MSE\\
         \cite{huang2022evaluation} & 2022 & FR & R & EKDL\\
         \cite{boutros2022template} & 2022 & PR & SRB, RB, FB, R & KLD, MSE, CS\\
         \cite{liu2022coupleface} & 2022 & FR & FB, R & N-$L_2$, CS\\
         \cite{boutros2022low} & 2022 & IR & FB & $L_2$ \\
         \cite{li2023rethinking} & 2023 & FR & RFB, FB & N-MSE \\
         \cite{zhao2023grouped} & 2023 & FR & RB & KLD \\
         \cite{xu2023probabilistic} & 2023 & FR & BEA & W$L_2$-MC\\
         \cite{aslam2023privileged} & 2023 & ER & R & CS \\
         \cite{caldeira2023unveiling} & 2023 & MAD & R & CS \\ 
    \end{tabular}
    \vspace{-1.5em}
    \label{tab:KD}
\end{table}
\subsection{Pruning}\label{sec:pruning_SOTA}

This section describes the current literature on pruning techniques in the fields of biometrics and CV. A summary of the analyzed studies can be found in Table~\ref{tab:pruning}. It is worth noting that for a greater focus on biometrics, the selected literature on computer vision represents some of the works responsible for driving the field toward further developments. 

\subsubsection{Pruning in Biometrics}

As quantization and KD, pruning techniques have also been used to compress models that process biometrics data. Polyak~\textit{et al.}~\cite{polyak2015channel} developed three LW pruning strategies for CNNs used in FR: inbound pruning (IP), reduce and reuse pruning (RRP) and hybrid pruning (HP). IP focuses on the relationship between the channels resulting from a certain layer and the filters to be applied in the next one. After determining a set of channels, the variance of their activation concerning each of the following filters is calculated. Then, the method prunes the channels that do not present a significant contribution to the information each filter extracts, meaning that, each filter will have a specific set of channels that are kept while others are disregarded, reducing the number of computations at inference time. This strategy follows an IPR approach and allows for some flexibility since the channels are selected by thresholding. Hence, the sparsity level is not predefined, meaning that filters whose number of insignificant connections is bigger will prune a bigger amount of channels. RRP uses a variance criterion to prune a specific number of channels from those that are outputted by a specific layer. This IPR method is based on the supposition that some of the channels encode redundant information that may be recovered by using the remaining channels to reconstruct the pruned ones. Finally, HP consists of applying RRP followed by IP. This is a viable option since RRP and IP have different targets inside the network and, as so, can be combined. Overall, HP performed better and IP proved to be more sensitive to pruning than RRP since the latter uses the remaining channels to reconstruct the pruned ones. Regarding the importance of following an IPR approach, Polyak~\textit{et al.}~\cite{polyak2015channel} showed that fine-tuning the model after pruning each layer is essential to reduce the performance drop. This is particularly true for deeper layers as the information they receive is affected by all the pruning errors accumulated from the previous layers. 

Alonso~\textit{et al.}~\cite{alonso2023squeezerfacenet} used a first-order Taylor approximation of the error induced in the loss when a filter is pruned to define the importance score of each filter. This method can be easily deployed by backpropagating the gradients of the loss, meaning that the pruning is similar to the training phase of a DL model. In each epoch, several mini-batches are passed through the model, allowing for the computation of the importance score of each filter. These scores are then averaged and used to prune a specific percentage of the network. The authors test frameworks where the comparison is made between pairs of faces or pairs of sets of five faces of the same identity since the usage of a bigger number of images to perform the comparison might help ensure that the most relevant features of each identity are retained while the random information is averaged out. The results obtained show that the validation accuracy reduces abruptly at the beginning of the pruning process but increases again until a certain level of sparsity is reached. This happens at about 15\% and 40\% sparsity when one and five images are used as a template, respectively, suggesting that using several input images makes the network stronger, enabling it to reach higher pruning sparsities. The retraining phase after each set of 5 epochs was useful to stabilize the network's performance and reduce its drop. The network's size only starts reducing significantly when about 10\% of the network is already pruned, showing that the first filters that are removed tend to be smaller or affect a lower number of channels than in later pruning stages. Besides, the final embedding vector is only reduced in size when about 18\% of the network is already pruned, proving that the last layer is not affected in the early pruning stages, as expected taking that this is the most specific layer for the task at hand. This goes in line with the conclusions withdrawn by Polyak~\textit{et al.}~\cite{polyak2015channel}. It could also be interesting to immediately eliminate the filters whose removal decreases the value of the loss, as will be further discussed in Section \ref{pruning_CV}.

Liu~\textit{et al.}~\cite{liu2021discrimination} presented a pruning strategy that effectively removes channels with a low contribution to the discriminative capacities of the network through the implementation of a loss function with two terms: a KD term that approximates the reconstructed feature maps to the ones generated by the original model and a term that considers the discriminative power of each channel or kernel through gradient evaluation. This methodology was tested in both IC and FR problems in an LW fashion, leading to better performances than the remaining evaluated techniques. The analysis of the feature maps obtained by pruned and not-pruned channels also showed that this methodology is successful in removing both the less informative and the redundant channels. As expected, the usage of an appropriate pruning criterion proved to be extremely important since random pruning results in degraded performance. Furthermore, larger models are easier to prune with small performance drops since they have more redundant connections. This strategy was also tested in CW and hybrid frameworks, with the latter consisting of LW pruning followed by CW pruning. The best-performing models were obtained using CW pruning, while the hybrid strategy only achieved marginal performance improvement when compared with LW pruning. As stated in the paper, this results from the fact that the hybrid strategy is severely conditioned by the LW pruning step, as starting by pruning entire channels removes kernels that might be considered useful if CW pruning was applied instead, resulting in suboptimal performance. To further study this problem, a reversed hybrid framework that performs CW pruning first could be tested.

Besides their usage in FR, pruning techniques have also been applied to other areas of biometrics, such as sclera segmentation (SS) \cite{vitek2023ipad}, facial expression recognition (FER) \cite{li2019graph} and face attribute classification (FAC) \cite{lin2022fairgrape}. The SS framework proposed by Vitek~\textit{et al.}~\cite{vitek2023ipad} works iteratively by pruning a specified percentage of the original number of FLOPs in each iteration and retraining before further pruning, allowing the model to adapt to its new architecture. This strategy differs from the traditional $L_1$ and $L_2$-norm pruning criteria as it does not focus on strongly activated filters. Instead, the authors define the most important filters as the ones whose activations are more distinct from the mean activation map as they are expected to contain a larger amount of new information. This novel loss term is balanced with the classical $L_1$ and $L_2$-norm losses, proving that a combination of both is generally better than using either of them separately. This study also helped to reinforce the importance of using an adequate pruning criterion since both the proposed methodology and the traditional metrics consistently outperform random pruning. Besides, the analysis of the pruned networks at different sparsity levels showed that models obtained in the initial phases of the pruning process tend to have better performances than the original network. This can be justified by the heavy overparametrization of the original networks, which can suffer a performance boost when useless filters are pruned. Furthermore, a visual comparison between the segmentation results achieved by the original model and its pruned versions revealed that pruning can result in the elimination of some false positives present in the original segmentation due to poor acquisition conditions, leading to a performance boost. Hence, the usage of pruning as a way of improving the performance of overparametrized networks should be further explored.

Li~\textit{et al.}~\cite{li2019graph} developed a pruning strategy for ensembles of FER models. Using big ensembles can lead to improved results since it ensures diversity among the used networks but ultimately leads to redundant classifiers. When a sample is fed to the ensemble, the algorithm determines which training samples are closer to it. Then, the subset of networks that perform better on these training samples is used to classify the test sample by majority voting. Although this method presents very good results when compared with other pruning strategies, it involves determining the similarity between each test sample and all the training samples to perform the neighborhood determination, which makes it hard to deploy in real-world scenarios, especially when using big training sets.

\subsubsection{Pruning in Computer Vision} \label{pruning_CV}

Apart from biometrics, pruning has also been applied to CV tasks. As previously shown by Wang~\textit{et al.}~\cite{wang2021teacher}, Hinton~\textit{et al.}~\cite{hinton2015distilling} and Aslam~\textit{et al.}~\cite{aslam2023privileged} for KD, Zhu~\textit{et al.}~\cite{zhu2017prune} also demonstrated that pruning a large model is better than training a smaller version from scratch, as pruned models consistently outperformed smaller networks with similar footprints in the tasks of image recognition, language modeling (LM) and sentence translation (ST). The conducted experiments also showed that large models suffer smaller drops in accuracy than smaller ones when pruned to the same sparsity percentage. However, pruning cannot be performed to an arbitrary level of sparsity without severe performance degradation, meaning that starting with a smaller model that does not need to be extremely sparsified to reach the end-goal size might be beneficial as long as it performs similarly to its larger alternative. 

Frankle~\textit{et al.}~\cite{frankle2018lottery} suggested an approach based on an interesting interpretation of pruning. When pruning a model, the resultant network can be interpreted as one of its multiple subnetworks. Since the original model is usually overparametrized, it will have redundant parameters that do not contribute significantly to improving its performance and, thus, their evolution during training can be considered irrelevant. As such, subnetworks that contain a large number of relevant parameters have the potential to exhibit very good results when trained from scratch since they do not have the setback of having to train a set of extra useless parameters. The authors adopt a pruning strategy that resembles the teacher assistant KD strategy proposed by Mirzadeh~\textit{et al.}~\cite{mirzadeh2020improved}. In this case, several iterations of SSPR are applied, resulting in successively sparser networks until the desired sparsity level is reached. Instead of retrained, however, the pruned network is trained from scratch by initializing its weights with the random values that were used to train the original network. This method allowed a Lenet trained in MNIST to be reduced to approximately 3.6\% of its original number of parameters before revealing any performance losses. However, the global training process is slow for very sparse networks since the usage of iterative pruning forces the model to be trained from scratch several times before achieving the desired compression rate. Besides, it is important to state that no comparison was established between training each obtained subnetwork from scratch or simply fine-tuning it. Despite not invalidating the results, the lack of depth in the performed analysis raises some doubts regarding the usefulness of this methodology.

A problem that arises when implementing pruning techniques is that generally they are applied independently in each layer or channel, both in SSPR and IPR schemes. Disentangling the pruning groups from each other might result in overseeing important connections that exist between them and that should be considered for efficient pruning of the whole network. This is particularly probable when dealing with low-level layers, whose pruning can have a big impact on the following layers. To address this issue, Yu~\textit{et al.}~\cite{yu2018nisp} presented a methodology (NISP) that considers information from the whole network to perform pruning, ensuring that the pruned parts contribute less to the model's predictive capabilities. Another interesting feature of this methodology is that it prunes neurons in fully connected (FC) layers and complete channels of neurons in convolutional layers, instead of individual connections. This is extremely important as it prevents the sparsification problem mentioned in Section~\ref{pruning_intro}, removing the need to use specific hardware to perform the computations. To decide which neurons should be pruned, an importance score is calculated for each feature outputted by the model's last layer using FB-KD. These scores are backpropagated to determine the importance of each neuron of the previous layers, meaning that the score of the early neurons is impacted by their contribution in deeper layers. Pruning is performed in an LW fashion, starting with the final layers. The model is only fine-tuned after all layers are pruned, resulting in a variant of the traditional SSPR technique. The comparison with SOTA methods showed that NISP can achieve lower drops in accuracy while leading to a bigger reduction in the total number of parameters and FLOPs of the model. Furthermore, the improvements are not only visible after fine-tuning the pruned model, but also before, proving that NISP is efficient in selecting which neurons should be pruned. Hence, this study presents an extremely interesting and differentiated approach that focuses on solving two of the biggest problems associated with pruning techniques, with visible success. It is suggested that future work directions include testing NISP within an altered IPR or hybrid framework that considers several fine-tuning steps instead of one, to see whether pruning-induced losses can be further reduced. 

Lee~\textit{et al.}~\cite{lee2018snip} also used backpropagation to determine the relevance of each connection. In this case, however, the goal was to prune a network without training it first, eliminating the computational cost of obtaining an original model with good performance. The importance of each connection is measured by evaluating the impact that ignoring it has on the final loss through gradient backpropagation. After pruning the connections that result in higher loss fluctuations, the model is trained for the first time. The proposed methodology has nearly invisible performance degradation for sparsity levels as high as 90\%. Even for 99\% sparsity, the model's error deviates less than 1 percentual point from the dense model. Hence, this simple and effective strategy leads to highly sparse models with negligible losses in accuracy. As a future development, it is suggested that the connections whose elimination decreases the loss of the model are immediately removed as this methodology preserves connections that lead to a small yet negative impact on the final loss value. This suggestion is inspired by Liu's~\textit{et al.}~\cite{liu2022coupleface} work, which removes harmful examples' contribution from the KD loss term.


Li~\textit{et al.}~\cite{li2016pruning} focused on pruning convolutional layers as they are usually responsible for the majority of the FLOPs of networks with both FC and convolutional layers, even when the latter represent a small percentage of the parameters. To avoid the computational overhead introduced by the utilization of sparse matrixes, entire filters are pruned, similar to Yu~\textit{et al.}~\cite{yu2018nisp}. Pruning entire filters is also advantageous in the reduction of the total number of parameters and FLOPs of the model since removing a filter reduces not only the number of operations within its layers but also of the following layer, which receives one less channel to process \cite{li2016pruning}. The importance scores are determined using the traditional $L_1$-norm and the final pruning is only performed after pruning each layer individually to different rates to determine their individual sensitivity to pruning. This way, a higher number of filters is preserved in the most sensitive layers, similar to Bunda~\textit{et al.}~\cite{bunda2022sub}. Although this induced an extra computational cost to the global pruning framework, it led to interesting conclusions that can be deployed in further studies, namely that deeper layers are generally harder to prune since they extract features that are more relevant to the task at hand, opposed to what was concluded by Bunda~\textit{et al.}~\cite{bunda2022sub} for quantization. Furthermore, the number of classes of the used dataset has a big impact on the sparsity level achievable without severe performance drops. This is not unexpected since models that deal with more classes need to extract a bigger amount of features from each sample to be able to distinguish between them. Following this premise, it would be interesting to address a new research direction focused on obtaining small networks for $n$-class classification by pruning large networks pre-trained with a $m$-class dataset on the same task, where $m>n$.

\begin{table*}[]
    \caption{Summary of the literature on pruning (the  selected fundamental works on computer vision have been the ones responsible for driving the topic further). It can be noted that a significant amount of studies followed v-SSPR strategies. \textbf{Granularity}: EP: ensemble pruning, H: hybrid; \textbf{Pruning Strategy}: DACP: discrimination-aware channel pruning, GDEP: graph-based dynamic ensemble pruning, v-SSPR: variant SSPR; \textbf{Pruning Criterion}: ADMAM: activation deviation from mean activation map, AV: activation variance, CDP: channels discriminative power, CIL: connection impact on the loss, EIL: error induced in the loss, FIS: feature-based importance score, GD: geodesic distance; ILDG: importance levels across demographic groups; *to reduce the pruning losses, the 90\% of sparsity does not affect the last layers, where 50\% of sparsity is used instead, **1 out of 799 classifiers, ***of the total number of FLOPs, ****experiments performed for each layer individually}
    \centering
    \begin{tabular}{c|c|c|c|c|c|c|c}
         \textbf{Area} & \textbf{Paper} & \textbf{Year} & \textbf{Task} & \textbf{Granularity} & \textbf{Pruning Strategy} & \textbf{Pruning Criterion} & \textbf{Highest Sparsity Level} \\
         \hline
         \multirow{5}{*}{Biometrics} & \cite{polyak2015channel} & 2015 & FR & LW & IPR & AV & 90\%*\\
         & \cite{li2019graph} & 2019 & FER & EP & GDEP & GD & $>$99\%**\\
         & \cite{liu2021discrimination} & 2021 & FR, IC & LW, CW, H & DACP & CDP & 70\%\\
         & \cite{alonso2023squeezerfacenet} & 2023 & FR & LW & v-SSPR & EIL & 75\%\\
         & \cite{vitek2023ipad} & 2023 & SS & CW &  v-SSPR & $L_1$, $L_2$, ADMAM & 75\%***\\
         \hline \hline
         \multirow{6}{*}{CV} & \cite{li2016pruning} & 2016 & IC & LW & SSPR & $L_1$ & 90\%****\\ 
         & \cite{zhu2017prune} & 2017 & IR, LM, ST & LW & v-SSPR & $L_1$ & 97.5\%\\
         & \cite{frankle2018lottery} & 2018 & IC & LW & v-SSPR & $L_1$ & $>$99\%\\
         & \cite{yu2018nisp} & 2018 & IC & LW & v-SSPR & FIS & 90\%\\
         & \cite{lee2018snip} & 2018 & IC & LW & SSPR & CIL & 99\% 
    \end{tabular}
    \vspace{-1.5em}
    \label{tab:pruning}
\end{table*}
\subsection{Literature Revisited} \label{sec:lit_revised}

Quantization techniques, although underdeveloped in biometrics, have evolved to the extent of allowing successful sub-byte quantization ($b<8$) \cite{bunda2022sub,kolf2022lightweight,zhou2017incremental,zhang2018lq}. While highly quantizing a network can result in severe performance degradation, the development of efficient QAT frameworks greatly mitigates these losses, resulting in highly compressed networks that are competitive with their FP versions \cite{bunda2022sub}. However, further evolution of these methods is still needed, to increase quantization efficiency for even lower values of $b$, especially in challenging tasks. Due to the high efficiency of QAT, the development of strategies that break the process into smaller blocks separated by retraining phases presents a promising research direction. In terms of conceptualization, this idea follows similar reasoning as the teacher-assistant framework proposed by Mirzadeh~\textit{et al.}~\cite{mirzadeh2020improved} for KD, as the usage of iterative quantization steps may allow for early compensation of the compression-induced losses, mitigating some of their effects before further compression. Similarly, the development of novel MPQ strategies is expected to lead to competitive results, as they allow for high flexibility while attending to the network's characteristics \cite{bunda2022sub}. The combination of these two directions within the same framework should also be considered, as it is equally important to ensure that the network is regaining the lost performance as to ensure that these drops are as low as possible. While MPQ can be seen as a way of learning how to better perform quantization considering the specificities of each network, further knowledge on how to perform quantization can be gained through the implementation of learnable quantizers. Although some works have already been presented in this area \cite{zhang2018lq,jeon2022mr,wang2022learnable}, there are still several possibilities to define new methods or mathematically refine the current ones, resulting in a promising research direction.

The current literature on KD proves the general superiority of FB-KD over RB-KD \cite{wang2021teacher,boutros2022template,kolf2023syper}. Although this is expected, the model's performance can also benefit from applying both methods simultaneously \cite{ge2020efficient}. The teacher-student gap also plays a major role in KD, representing one of the main bottlenecks for its success. Although enlarging this gap leads to higher compression rates, it can highly undermine the student's understanding of the teacher's knowledge, resulting in inefficient distillation. A way to address this issue is adapting the teacher's knowledge to a simpler domain through the usage of cross-dataset distillation when privacy concerns are present \cite{ge2020efficient}. It is also possible to perform KD iteratively by adding teacher-assistants with intermediate complexity levels to the framework, as suggested by Mirzadeh~\textit{et al.}~\cite{mirzadeh2020improved}. In this scenario, the pretrained teacher transfers knowledge to the more complex assistant which then acts as a teacher for the next more complex assistant and so forth until the student model is reached. This prevents the student from accessing teacher information that it cannot fully understand, refocusing on the filtered features that are passed through the assistants and, thus, mitigating the teacher-student gap. To the extent of our knowledge, this technique is yet to be deployed in the field of biometrics, corroborating that methodologies that focus on reducing the teacher-student gap should be studied in deeper detail. Furthermore, the strategy proposed by Mirzadeh~\textit{et al.}~\cite{mirzadeh2020improved} presents a way of pursuing another research direction: the selection of an appropriate teacher for a specific student architecture. This task is not straightforward and, to the extent of our knowledge, it is still to be properly addressed by the DL community. The utilization of NAS, as done by Wang~\textit{et al.}~\cite{wang2021teacher}, is also a viable option to explore this topic.

Regarding other future developments, most of the studies conducted so far show that introducing selectivity to the KD procedure can highly benefit the final model's performance. On one side, avoiding the distillation of information regarding samples whose label is incorrectly predicted by the teacher has proved to result in enhanced performance \cite{aslam2023privileged,ge2018low}. On the other side, teacher models are not flawless and, thus, blindly transferring their knowledge to the student networks is not always advantageous \cite{zhao2023grouped,liu2022coupleface}. Techniques that prevent incorrect knowledge from being transferred are usually easy to interpret and deploy, resulting in more competitive students. Besides, forcing the student to exactly mimic its teacher's behavior can lead to over-regularization, resulting in downgraded performance. While some R-KD techniques that avoid this problem by selecting a less constrained knowledge to be distilled \cite{caldeira2023unveiling,duong2019shrinkteanet,aslam2023privileged} have already been developed, they are not yet widely deployed, presenting a research direction that should be further pursued by the DL community. 

While quantization and pruning strategies are mainly designed to perform compression, KD can be used without altering the model's architecture, by distilling knowledge to a network that works on harder ~\cite{huber2021mask} or lower-quality \cite{ge2018low,ge2020efficient,boutros2022low} data, thus performing a more complex task. The broader spectrum of KD applications helps explain the gap in biometrics literature for both quantization and pruning in comparison with KD. More research is needed to support the development of quantization and pruning strategies while enabling a fairer comparison between their and KD's contributions.

The analyzed pruning methodologies have consistently proved the efficiency of this technique in removing both the less informative and the redundant parts of the network \cite{liu2021discrimination,lee2018snip}. When the original model is highly overparametrized, pruning it to a small extent can even boost its performance, due to the elimination of useless connections \cite{vitek2023ipad}. Although pruning is mostly used as a compression tool, these findings suggest that it can enhance the efficiency of overparametrized models, revealing a promising research direction. Despite the high compression power of these techniques, there is an urgent need to address the problem of network sparsification through the implementation of pruning strategies that remove entire layers or channels \cite{li2016pruning,yu2018nisp}. 

Most of the withdrawn conclusions regarding pruning are analogous to the ones mentioned for quantization. On one side, the presence of retraining steps between the pruning stages can benefit the compressed model's performance \cite{frankle2018lottery,alonso2023squeezerfacenet,vitek2023ipad}. However, when the outputs of a layer or channel are little affected by the pruning process, immediate retraining might be unnecessary, as SSPR works similarly to IPR, without the extra computational burden. This highlights the need to redefine SSPR strategies and introduce iterative SSPR frameworks by grouping layers or channels into sub-networks that present low sensitivity to pruning. These sub-networks can then be internally pruned using SSPR while the whole network still follows an IPR framework where each sub-network is interpreted as an independent pruning group. On the other side, pruning strategies that focus on understanding which connections have a lower impact on the model's performance have achieved competitive results, proving to be advantageous against the traditional pruning criteria \cite{yu2018nisp,lee2018snip}. Opposed to quantization \cite{bunda2022sub}, however, pruning seems to have a more negative impact when performed in the last layers of the model \cite{polyak2015channel,alonso2023squeezerfacenet,li2016pruning}, which are more specific to the model's task, highlighting the disparities between different compression methods. 

The disparities between the evolutionary state of quantization, KD and pruning do not prevent several analogies from being established between them. Apart from the superiority of compressed models when compared with similar architectures trained from scratch, these techniques have proved to be more efficient when logical and deterministic frameworks are developed to perform compression. Whether by giving preference to compressing model parts more robust to compression-induced losses \cite{bunda2022sub,liu2021discrimination,vitek2023ipad} or by seeking mathematical logic when designing a KD framework \cite{caldeira2023unveiling}, careful formulation of the developed strategy is essential to exploit these techniques to its full potential. As such, this area of DL can highly benefit from the study of novel ways to perform compression, instead of directly replicating traditional methodologies. 

The impact of the complexity difference between the model and its compressed version on the final performance is another common point between the analyzed methodologies. Whether measured by the value of $b$ (quantization), the teacher-student gap (KD) or the sparsity level (pruning), larger complexity gaps result in higher compression-induced performance drops. However, it is not trivial to directly deploy the knowledge gathered from previous studies in a novel field, as different frameworks present distinct robustness to compression-induced losses \cite{bunda2022sub,kolf2022lightweight}. Hence, instead of singularly testing a compression framework, studying the impact that the above-mentioned factors specific to each compression technique have on performance is of the utmost importance.

It is also important to highlight the complementary nature of the analyzed compression techniques. Quantization, KD and pruning produce distinct effects on models, making the combination of two or more compression techniques possible within the same compression framework \cite{ji2022neural}. However, the broad spectrum of applications of KD allows its combination with quantization and pruning in distinct ways. KD has been used in the determination of connections' importance scores in pruning frameworks \cite{liu2021discrimination,yu2018nisp} but can also be seen as a fine-tuning tool of the compressed model \cite{boutros2022quantface}, while simultaneously being useful in the reduction of compression-induced bias \cite{blakeney2021simon}, as will be discussed in Section \ref{sec:adv_vs_disadv}. This demonstrates the multipurpose of KD and emphasizes the interest in combining this tactic with other compression methods.
\section{Hidden Implications of Compression on Bias}\label{sec:adv_vs_disadv}

Compression methods have proved to highly reduce models' computational and memory costs, without significant performance degradation. From this point of view, compression seems to be an almost flawless methodology to adapt heavy models to scenarios with limited resources. However, it should be noted that a significant amount of studies conducted in this area only focus on evaluating the performance degradation quantitatively, lacking a deeper analysis of its origin. In most cases, the global losses are quantified as small and immediately neglected based on the assumption that this performance loss is uniformly distributed over the data, which might not be true. The model might be ignoring underrepresented groups for the sake of global performance when compression is performed, meaning that these subsets' samples will be misclassified more often than the remaining ones, resulting in increased bias.


Different definitions of bias were brought to the surface in DL research. However, most sources agree that it relates to performance differences that are influenced by a particular sub-population \cite{robinson2020face}, at least in biometrics. Several studies showed that the recognition performance using FR systems varies based on gender \cite{DBLP:conf/bmvc/AlbieroB20, DBLP:conf/icb/AlbieroZB20,DBLP:conf/icb/FuD22}, age \cite{DBLP:conf/icb/DebN018} or ethnicity \cite{DBLP:journals/pami/HuangLLT20}. In Terhorst's~\textit{et al.}~\cite{terhorst2020compr} comprehensive study, the investigation of performance differences has been extended to consider non-demographic attributes such as expression, pose or illumination. Bias is not limited to biometric applications in the area of FR but is also present in other biometric tasks such as presentation attack detection \cite{fang2024fairpad,DBLP:conf/eusipco/FangDKK20}, face image quality assessment \cite{DBLP:conf/icb/TerhorstKDKK20}, face detection \cite{DBLP:conf/fgr/MittalTMVS23}, or even explainability tools \cite{DBLP:conf/eusipco/HuberFBD23}. Bias is also commonly present in human observers, who tend to correctly classify more faces presenting demographic characteristics they contact with often \cite{robinson2020face}.

To perform an analysis regarding the performance differences, it is necessary to provide a suitable test framework that allows a comparison based on the different investigated sub-populations. Wang~\textit{et al.}~\cite{wang2019racial} proposed the Racial Faces in the Wild (RFW) dataset, which contains equal percentages of faces from Caucasian, Indian, Asian and African identities. They proposed an ethnicity-balanced training set to reduce the ethnicity bias, but models trained on RFW still perform better for Caucasian faces \cite{wang2019racial}. A similar approach was proposed by Karkkainen~\textit{et al.}~\cite{karkkainen2021fairface}, which proposed FairFace, a race-balanced dataset that is diverse in terms of race, age, expressions, head orientation, and photographic conditions. Robinson~\textit{et al.}~\cite{robinson2020face} also proposed another balanced dataset in terms of race, gender and both gender and race, Balanced Faces in the wild (BFW). A different understanding of balance has been proposed by Wang~\textit{et al.}~\cite{wang2021meta}. In contrast to other works, their two novel datasets either follow the real-world skin-tone distribution (BUPT-Globalface) or are balanced in equal proportions among all the considered skin-tones (BUPT-Balancedface). However, contrary to what the authors suggest, BUPT-Globalface and BUPT-Balancedface do not follow a similar distribution across different skin tones for gender. As an example, Tone-III and Tone-VIII have about 45\% and 5\% of females respectively, which might induce gender bias. Although complete balance is not easy to achieve, benchmarks that are well-balanced regarding several sensitive attributes and not only one should be developed. Ignoring this need might result in models that are unbiased regarding the characteristic they are being analyzed for (race, in this case) while withholding increased bias regarding other sensitive attributes (such as gender), as will be discussed later in this section.  

Besides more balanced datasets to mitigate bias, different approaches have been proposed to increase the fairness of biometric models. Wang~\textit{et al.}~\cite{wang2021meta} evaluated the performance differences from the different complexity levels of the analyzed demographic sub-groups. They considered different margins for the evaluated skin tones and used an extra-balanced dataset to update the margins of the darker skin categories during training to adjust to the different complexity levels. The adaptive margins result in a better alignment of the model's feature space with the real data distribution, as harder skin tones can be assigned larger margins. Xu~\textit{et al.}~\cite{xu2020investigating} aimed at mitigating bias in facial ER by either removing the sub-group attribute information or highlighting it during their recognition process. The least bias was achieved when the amount of sensitive information in the logits was minimized, proving that eliminating sensitive attributes before classification might result in decreased performance differences.


As such, bias is a major problem in human-related data processing that is receiving increasing attention and that requires special awareness during the development of biometrics DL models. This also applies to compressed models, whose reduction in size might impact model performance differently for distinct demographic groups. Some works that address this question have already been conducted not only in the biometrics area \cite{stoychev2022effect,neto2023compressed, DBLP:conf/iccvw/LiuZSYL21,blakeney2021simon,DBLP:conf/cvpr/IofinovaPA23,lin2022fairgrape} but also in other fields such as NLP \cite{ahn-etal-2022-knowledge,goncalves-strubell-2023-understanding}.

Stoychev~\textit{et al.}~\cite{stoychev2022effect} analyzed the impact of quantization and pruning on the gender bias of FER models. The applied PTQ to 8-bit precision resulted in a final model with similar accuracy and fairness as the original one. Pruning, on the other hand, increased the disparities between men's and women's accuracy, resulting in decreased fairness. With CK+, this was verified for pruning percentages as low as 10\%, where the accuracy gap between genders increased from 2.36 (baseline) to 15.60 percentual points, even though low global accuracy drops were verified. This was not the case when considering a baseline with weights close to 0 (RAF) since pruning them had a reduced effect on the overall model’s performance. While the global performance of the CK+ baseline suffered considerable drops at about 60\% sparsity, this only happened at about 90\% sparsity for RAF. The obtained results are highly dependent on the selected compression strategies, meaning that they should not be taken as absolute. Without performing further experiments it is impossible to know, for instance, whether the usage of a lower bit precision for quantization would result in increased bias. 

Neto~\textit{et al.}~\cite{neto2023compressed} analyzed whether quantization resulted in increased racial bias while testing the impact that using different types of data to perform the compression has on this bias. Using the models available in Boutros~\textit{et al.}~\cite{boutros2022quantface} and others trained on racially-balanced datasets, the authors test them in the well-balanced datasets proposed by Wang~\textit{et al.}~\cite{wang2019racial} and Wang~\textit{et al.}~\cite{wang2021meta}, and a dataset of synthetic data, to verify whether using data that does not rise privacy concerns has any impact in the racial bias of the compressed models. The obtained results confirmed that smaller models achieve lower performance and fairness. Models quantized with synthetic data presented reduced bias than models quantized with real data. A possible justification is that using data with different distributions to train and quantize the model might be useful to shield it against racial bias. This assumption also goes in line with the conclusions withdrawn for models trained and quantized with different real data, since they tend to perform better than the ones trained and quantized with the same data. Hence, it could be interesting to use both real and synthesized data in pruning and KD frameworks and to test whether this impacts other types of bias, as suggested by the authors.

Liu~\textit{et al.}~\cite{DBLP:conf/iccvw/LiuZSYL21} proposed a strategy to rectify the data bias in KD procedures to achieve fairer models with increased performance. When knowledge is distilled to a student, the distribution of its training data should be representative of the teacher space to ensure that its relevant characteristics are distilled. Ideally, this dataset's features should be distributed as uniformly as possible in the teacher space. Hence, two strategies that extract a more uniform subsample of the selected training set are evaluated. These strategies outperformed simple KD in the evaluated scenarios, resulting in a significant reduction of class imbalance. Furthermore, samples with a higher level of difficulty (such as blurry pictures) are sampled more often than images that are easier to analyze (such as front-view pictures), balancing the bias induced by the distinct complexity levels of the considered samples. Even when the student is trained from scratch, the usage of the presented strategies to sample an appropriate training dataset results in increased performance compared with simple KD without sampling, suggesting that data balance is of the utmost importance to increase fairness and boost performance.

Blakeney~\textit{et al.}~\cite{blakeney2021simon} verified whether using KD to fine-tune the pruned network helped mitigate pruning-induced bias. The pruning criterion aimed to balance the network's FNR/FPR ratio, ensuring that it remained close to the one presented by the original model. Furthermore, model pruning was performed iteratively, facilitating the recovery from pruning-induced losses. Despite these efforts, pruning still led to increased bias. Further training with KD partially eliminated this bias, resulting in models with better-balanced changes of FNR and FPR. Besides, methods that considered not only RB-KD but also FB-KD were more effective in reducing the induced bias, which is congruent with the discussion presented in Section \ref{sec:KD_SOTA}. Hence, this study shows that using KD on top of other techniques might reduce compression-induced bias. 

Paganini~\textit{et al.}~\cite{paganini2020prune} focused on the analysis of pruning-induced bias. Their experiments show that performance degradation is dependent on several factors, such as the initial performance and the desired sparsity. It also depends on class imbalance and complexity, since complex and/or underrepresented classes achieve lower performances, resulting in bias. As an example, a study on a version of MNIST showed that the digits' angle of tilt and thickness influenced the performance degradation after pruning which aligns with the fact that samples with different complexities can induce bias even in well-balanced datasets \cite{wang2019racial,xu2020investigating}. To address this problem, it would be interesting to test whether balancing the trainset by considering the relative complexity of each group could help mitigate the induced bias \cite{DBLP:conf/iccvw/LiuZSYL21}. In this problem, the usage of a dataset with a bigger amount of samples whose digits are tilted could even out the bias induced by their higher complexity. An alternative would be to maintain a well-balanced dataset and use a weighted loss, giving higher importance to more complex groups. Apart from the need for a metric to evaluate the complexity of each group of samples, these groups also need to be properly defined and balanced (or their weights), which can be hard when several characteristics that induce complexity are considered. Ignoring some of them might compromise bias mitigation while acknowledging them all might lead to undesired scenarios brought by extreme specificity, such as each sample being considered a separate group. 

Iofinova~\textit{et al.}~\cite{DBLP:conf/cvpr/IofinovaPA23} analyzed the impact of pruning on model bias by studying versions of the same dense model pruned to sparsity levels between 80\% and 99.5\%. The fairness evaluation was performed by measuring the bias amplification, that is the extent to which the model amplifies the correlation between the predicted attributes and specific identity information (such as gender and age). Hence, this metric reveals how stereotyped are the models' predictions. The considered models were trained to identify facial binary attributes in both single (only one attribute is predicted) and joint (more than one attribute is predicted) training schemes. While pruning up to 95\% sparsity did not result in a significant decrease in fairness, the bias increased significantly for higher sparsity values, showing a strong correlation with the bias of their dense version. Models with joint-train frameworks presented lower bias for lower sparsity values since a more complete evaluation of facial characteristics leads to a stronger feature representation that discourages bias. At higher sparsities, however, the bias increases as the compactness effect surpasses the initial robustness of the model. The authors also analyzed how to mitigate compression-induced bias, proving that both threshold calibration and the override of sensitive samples can contribute to largely decreasing the bias amplification, even for high sparsity levels.

Lin~\textit{et al.}~\cite{lin2022fairgrape} proposed FairGRAPE, a pruning method that highlights the importance of developing fairness-aware compression techniques in biometrics. To mitigate pruning-induced bias, the pruning criterion should be adapted to evaluate the importance of each connection of the network individually for a set of predefined groups, instead of considering the dataset as a whole. By balancing the importance levels across different demographic groups, it is possible to reduce the variance of performance differences between them and achieve a fairer model. The pruning strategy is similar to the one presented by Frankle~\textit{et al.}~\cite{frankle2018lottery}. On the classification of sensitive attributes, both gender and race classifiers were able to reach good results. FairGRAPE not only achieved better overall performance when compared with SOTA methods but also proved to be fairer, as it resulted in lower STD of accuracy and loss across the analyzed sensitive groups. Although the verified overall performance drops are acceptable taking into consideration the high pruning sparsity levels that were tested (from 90\% to 99\%), there is still a large margin for improvement regarding model bias. As an example, in the race classification task on the FairFace dataset, the FairGRAPE model presents an accuracy of 80.3\% in Black identities and only 47.5\% in Hispanic ones. Besides, the ratio of importance scores of the original model for different groups is defined as the target for its pruned version. This can be restraining since the original model sets the upper boundary of the pruned model's fairness. Hence, this technique should be tested using the same importance scores for every predefined group or scores that consider each group's complexity \cite{wang2019racial,xu2020investigating,wang2021meta}, especially when the original model is severely biased.
\section{Conclusion} \label{sec:conclusion}





The present document covered the compression works available in the biometrics field while considering the available literature on computer vision. Apart from the mathematical definition of quantization, knowledge distillation and pruning (Section \ref{sec:compression}), this survey presented an extensive analysis of the current compression literature and distinct usages for traditional compression methods (Section \ref{sec:sota}). While KD methodologies have been widely deployed in the biometrics field, quantization and pruning are still to be broadly implemented in this area, requiring further study not only to boost their development but also to allow for a fairer comparison between the usefulness of the analyzed compression methods. Section \ref{sec:sota} also discussed the advantages and disadvantages of the proposed techniques, while suggesting future research directions on compression based on the current SOTA.

The analysis of the available literature showed that model bias is still a problem in the field of biometrics (Section \ref{sec:adv_vs_disadv}), where it can result in particularly harmful situations such as the discrimination of underrepresented demographic groups. While balanced datasets regarding characteristics such as race and gender are beneficial, their development is challenging, especially when several demographics need to be balanced. Furthermore, compression is often responsible for amplifying model bias. This can be attributed to a larger loss in accuracy in the underrepresented groups to mitigate the global compression-induced performance drop. Hence, the development of fair compression strategies is of the utmost importance and should be carefully addressed. As such, we propose a set of interesting future research directions on this topic, thus contributing to the evolution towards fairer compression in biometrics.


 \bibliographystyle{ieeetr} 
 
 \bibliography{refs}

\begin{thebibliography}{10}

\bibitem{neto2022imil4path}
P.~C. Neto, S.~P. Oliveira, D.~Montezuma, J.~Fraga, A.~Monteiro, L.~Ribeiro,
  S.~Gon{\c{c}}alves, I.~M. Pinto, and J.~S. Cardoso, ``imil4path: A
  semi-supervised interpretable approach for colorectal whole-slide images,''
  {\em Cancers}, vol.~14, no.~10, p.~2489, 2022.

\bibitem{melo2023synthesis}
T.~Melo, J.~Cardoso, A.~Carneiro, A.~Campilho, and A.~M. Mendonca, ``Oct image
  synthesis through deep generative models,'' in {\em CBMS}, pp.~561--566,
  2023.

\bibitem{kolf2022lightweight}
J.~N. Kolf, F.~Boutros, F.~Kirchbuchner, and N.~Damer, ``Lightweight periocular
  recognition through low-bit quantization,'' in {\em IJCB}, pp.~1--12, IEEE,
  2022.

\bibitem{vitek2023ipad}
M.~Vitek, M.~Bizjak, P.~Peer, and V.~{\v{S}}truc, ``Ipad: Iterative pruning
  with activation deviation for sclera biometrics,'' {\em Journal of King Saud
  University-Computer and Information Sciences}, vol.~35, no.~8, p.~101630,
  2023.

\bibitem{neto2022explainable}
P.~C. Neto, T.~Gon{\c{c}}alves, J.~R. Pinto, W.~Silva, A.~F. Sequeira, A.~Ross,
  and J.~S. Cardoso, ``Explainable biometrics in the age of deep learning,''
  {\em arXiv preprint arXiv:2208.09500}, 2022.

\bibitem{delgado2023m}
P.~Delgado-Santos, R.~Tolosana, R.~Guest, R.~Vera-Rodriguez, and J.~Fierrez,
  ``M-gaitformer: Mobile biometric gait verification using transformers,'' {\em
  Engineering Applications of Artificial Intelligence}, vol.~125, p.~106682,
  2023.

\bibitem{kocacinar2022real}
B.~Kocacinar, B.~Tas, F.~P. Akbulut, C.~Catal, and D.~Mishra, ``A real-time
  cnn-based lightweight mobile masked face recognition system,'' {\em Ieee
  Access}, vol.~10, pp.~63496--63507, 2022.

\bibitem{boutros2020benchmarking}
F.~Boutros, N.~Damer, K.~Raja, R.~Ramachandra, F.~Kirchbuchner, and A.~Kuijper,
  ``On benchmarking iris recognition within a head-mounted display for ar/vr
  applications,'' in {\em IJCB}, pp.~1--10, IEEE, 2020.

\bibitem{DBLP:journals/ivc/BoutrosDRRKK20}
F.~Boutros, N.~Damer, K.~B. Raja, R.~Ramachandra, F.~Kirchbuchner, and
  A.~Kuijper, ``Iris and periocular biometrics for head mounted displays:
  Segmentation, recognition, and synthetic data generation,'' {\em Image Vis.
  Comput.}, vol.~104, p.~104007, 2020.

\bibitem{miller2022temporal}
R.~Miller, N.~K. Banerjee, and S.~Banerjee, ``Temporal effects in motion
  behavior for virtual reality {(VR)} biometrics,'' in {\em {VR}},
  pp.~563--572, {IEEE}, 2022.

\bibitem{gou2021knowledge}
J.~Gou, B.~Yu, S.~J. Maybank, and D.~Tao, ``Knowledge distillation: A survey,''
  {\em Int. J. Comput. Vis.}, vol.~129, pp.~1789--1819, 2021.

\bibitem{ge2018low}
S.~Ge, S.~Zhao, C.~Li, and J.~Li, ``Low-resolution face recognition in the wild
  via selective knowledge distillation,'' {\em {IEEE} Trans. Image Process.},
  vol.~28, no.~4, pp.~2051--2062, 2018.

\bibitem{krishnamoorthi2018quantizing}
R.~Krishnamoorthi, ``Quantizing deep convolutional networks for efficient
  inference: A whitepaper,'' {\em arXiv preprint arXiv:1806.08342}, 2018.

\bibitem{gholami2022survey}
A.~Gholami, S.~Kim, Z.~Dong, Z.~Yao, M.~W. Mahoney, and K.~Keutzer, ``A survey
  of quantization methods for efficient neural network inference,'' in {\em
  Low-Power Computer Vision}, pp.~291--326, Chapman and Hall/CRC, 2022.

\bibitem{zhu2017prune}
M.~Zhu and S.~Gupta, ``To prune, or not to prune: Exploring the efficacy of
  pruning for model compression,'' 2018.

\bibitem{wang2021teacher}
X.~Wang, ``Teacher guided neural architecture search for face recognition,'' in
  {\em AAAI}, vol.~35, pp.~2817--2825, 2021.

\bibitem{neto2023compressed}
P.~C. Neto, E.~Caldeira, J.~S. Cardoso, and A.~F. Sequeira, ``Compressed models
  decompress race biases: What quantized models forget for fair face
  recognition,'' in {\em BIOSIG}, pp.~1--5, IEEE, 2023.

\bibitem{boutros2022quantface}
F.~Boutros, N.~Damer, and A.~Kuijper, ``Quantface: Towards lightweight face
  recognition by synthetic data low-bit quantization,'' in {\em ICPR},
  pp.~855--862, IEEE, 2022.

\bibitem{choi2020data}
Y.~Choi, J.~Choi, M.~El-Khamy, and J.~Lee, ``Data-free network quantization
  with adversarial knowledge distillation,'' in {\em CVPR Workshops},
  pp.~710--711, 2020.

\bibitem{zhou2017incremental}
A.~Zhou, A.~Yao, Y.~Guo, L.~Xu, and Y.~Chen, ``Incremental network
  quantization: Towards lossless {CNN}s with low-precision weights,'' in {\em
  ICLR}, 2017.

\bibitem{jacob2018quantization}
B.~Jacob, S.~Kligys, B.~Chen, M.~Zhu, M.~Tang, A.~Howard, H.~Adam, and
  D.~Kalenichenko, ``Quantization and training of neural networks for efficient
  integer-arithmetic-only inference,'' in {\em CVPR}, pp.~2704--2713, 2018.

\bibitem{li2016pruning}
H.~Li, A.~Kadav, I.~Durdanovic, H.~Samet, and H.~P. Graf, ``Pruning filters for
  efficient convnets,'' in {\em ICLR}, 2017.

\bibitem{luo2016face}
P.~Luo, Z.~Zhu, Z.~Liu, X.~Wang, and X.~Tang, ``Face model compression by
  distilling knowledge from neurons,'' in {\em AAAI}, vol.~30, 2016.

\bibitem{polyak2015channel}
A.~Polyak and L.~Wolf, ``Channel-level acceleration of deep face
  representations,'' {\em IEEE Access}, vol.~3, pp.~2163--2175, 2015.

\bibitem{neto2022myope}
P.~C. Neto, A.~F. Sequeira, and J.~S. Cardoso, ``Myope models-are face
  presentation attack detection models short-sighted?,'' in {\em WACV
  Workshops}, pp.~390--399, 2022.

\bibitem{boutros2022pocketnet}
F.~Boutros, P.~Siebke, M.~Klemt, N.~Damer, F.~Kirchbuchner, and A.~Kuijper,
  ``Pocketnet: Extreme lightweight face recognition network using neural
  architecture search and multistep knowledge distillation,'' {\em IEEE
  Access}, vol.~10, pp.~46823--46833, 2022.

\bibitem{wang2022learnable}
L.~Wang, X.~Dong, Y.~Wang, L.~Liu, W.~An, and Y.~Guo, ``Learnable lookup table
  for neural network quantization,'' in {\em IEEE/CVF CVPR}, pp.~12423--12433,
  2022.

\bibitem{pytorch2019}
A.~Paszke, S.~Gross, F.~Massa, A.~Lerer, J.~Bradbury, G.~Chanan, T.~Killeen,
  Z.~Lin, N.~Gimelshein, L.~Antiga, A.~Desmaison, A.~Kopf, E.~Yang, Z.~DeVito,
  M.~Raison, A.~Tejani, S.~Chilamkurthy, B.~Steiner, L.~Fang, J.~Bai, and
  S.~Chintala, ``Pytorch: An imperative style, high-performance deep learning
  library,'' in {\em NeurIPS}, pp.~8024--8035, 2019.

\bibitem{miyashita2016convolutional}
D.~Miyashita, E.~H. Lee, and B.~Murmann, ``Convolutional neural networks using
  logarithmic data representation,'' {\em arXiv preprint arXiv:1603.01025},
  2016.

\bibitem{zhang2018lq}
D.~Zhang, J.~Yang, D.~Ye, and G.~Hua, ``Lq-nets: Learned quantization for
  highly accurate and compact deep neural networks,'' in {\em ECCV},
  pp.~365--382, 2018.

\bibitem{jeon2022mr}
Y.~Jeon, C.~Lee, E.~Cho, and Y.~Ro, ``Mr. biq: Post-training non-uniform
  quantization based on minimizing the reconstruction error,'' in {\em CVPR},
  pp.~12329--12338, 2022.

\bibitem{nair2010rectified}
V.~Nair and G.~E. Hinton, ``Rectified linear units improve restricted boltzmann
  machines,'' in {\em ICML}, pp.~807--814, 2010.

\bibitem{bunda2022sub}
S.~Bunda, L.~Spreeuwers, and C.~Zeinstra, ``Sub-byte quantization of mobile
  face recognition convolutional neural networks,'' in {\em BIOSIG}, pp.~1--5,
  IEEE, 2022.

\bibitem{hinton2015distilling}
G.~Hinton, O.~Vinyals, and J.~Dean, ``Distilling the knowledge in a neural
  network,'' {\em arXiv preprint arXiv:1503.02531}, 2015.

\bibitem{aslam2023privileged}
M.~H. Aslam, M.~O. Zeeshan, M.~Pedersoli, A.~L. Koerich, S.~Bacon, and
  E.~Granger, ``Privileged knowledge distillation for dimensional emotion
  recognition in the wild,'' in {\em CVPRW}, pp.~3337--3346, 2023.

\bibitem{yu2018nisp}
R.~Yu, A.~Li, C.-F. Chen, J.-H. Lai, V.~I. Morariu, X.~Han, M.~Gao, C.-Y. Lin,
  and L.~S. Davis, ``Nisp: Pruning networks using neuron importance score
  propagation,'' in {\em CVPR}, pp.~9194--9203, 2018.

\bibitem{huber2021mask}
M.~Huber, F.~Boutros, F.~Kirchbuchner, and N.~Damer, ``Mask-invariant face
  recognition through template-level knowledge distillation,'' in {\em FG},
  pp.~1--8, IEEE, 2021.

\bibitem{ge2020efficient}
S.~Ge, S.~Zhao, C.~Li, Y.~Zhang, and J.~Li, ``Efficient low-resolution face
  recognition via bridge distillation,'' {\em {IEEE} Trans. Image Process.},
  vol.~29, pp.~6898--6908, 2020.

\bibitem{boutros2022low}
F.~Boutros, O.~Kaehm, M.~Fang, F.~Kirchbuchner, N.~Damer, and A.~Kuijper,
  ``Low-resolution iris recognition via knowledge transfer,'' in {\em BIOSIG},
  pp.~1--5, IEEE, 2022.

\bibitem{zhao2023grouped}
W.~Zhao, X.~Zhu, K.~Guo, X.~Zhang, and Z.~Lei, ``Grouped knowledge distillation
  for deep face recognition,'' pp.~3615--3623, 2023.

\bibitem{kolf2023syper}
J.~N. Kolf, J.~Elliesen, F.~Boutros, H.~Proen{\c{c}}a, and N.~Damer, ``Syper:
  Synthetic periocular data for quantized light-weight recognition in the nir
  and visible domains,'' {\em Image Vis. Comput.}, vol.~135, p.~104692, 2023.

\bibitem{wang2021knowledge}
L.~Wang and K.-J. Yoon, ``Knowledge distillation and student-teacher learning
  for visual intelligence: A review and new outlooks,'' {\em {IEEE} Trans.
  Pattern Anal. Mach. Intell.}, vol.~44, no.~6, pp.~3048--3068, 2021.

\bibitem{duong2019shrinkteanet}
C.~N. Duong, K.~Luu, K.~G. Quach, and N.~Le, ``Shrinkteanet: Million-scale
  lightweight face recognition via shrinking teacher-student networks,'' {\em
  arXiv preprint arXiv:1905.10620}, 2019.

\bibitem{wu2020learning}
X.~Wu, R.~He, Y.~Hu, and Z.~Sun, ``Learning an evolutionary embedding via
  massive knowledge distillation,'' {\em International Journal of Computer
  Vision}, vol.~128, pp.~2089--2106, 2020.

\bibitem{liu2022coupleface}
J.~Liu, H.~Qin, Y.~Wu, J.~Guo, D.~Liang, and K.~Xu, ``Coupleface: Relation
  matters for face recognition distillation,'' in {\em ECCV}, pp.~683--700,
  Springer, 2022.

\bibitem{li2023rethinking}
J.~Li, Z.~Guo, H.~Li, S.~Han, J.-w. Baek, M.~Yang, R.~Yang, and S.~Suh,
  ``Rethinking feature-based knowledge distillation for face recognition,'' in
  {\em CVPR}, pp.~20156--20165, 2023.

\bibitem{boutros2022template}
F.~Boutros, N.~Damer, K.~Raja, F.~Kirchbuchner, and A.~Kuijper,
  ``Template-driven knowledge distillation for compact and accurate periocular
  biometrics deep-learning models,'' {\em Sensors}, vol.~22, no.~5, p.~1921,
  2022.

\bibitem{huang2022evaluation}
Y.~Huang, J.~Wu, X.~Xu, and S.~Ding, ``Evaluation-oriented knowledge
  distillation for deep face recognition,'' in {\em CVPR}, pp.~18740--18749,
  2022.

\bibitem{caldeira2023unveiling}
E.~Caldeira, P.~C. Neto, T.~Goncalves, N.~Damer, A.~F. Sequeira, and J.~S.
  Cardoso, ``Unveiling the two-faced truth: Disentangling morphed identities
  for face morphing detection,'' in {\em EUSIPCO}, 2023.

\bibitem{boutros2020compact}
F.~Boutros, N.~Damer, M.~Fang, K.~Raja, F.~Kirchbuchner, and A.~Kuijper,
  ``Compact models for periocular verification through knowledge
  distillation,'' in {\em BIOSIG}, pp.~1--5, IEEE, 2020.

\bibitem{chechik1998synaptic}
G.~Chechik, I.~Meilijson, and E.~Ruppin, ``Synaptic pruning in development: a
  computational account,'' {\em Neural computation}, vol.~10, no.~7,
  pp.~1759--1777, 1998.

\bibitem{zukerman2011brain}
W.~Zukerman and A.~Purcell, ``Brain's synaptic pruning continues into your
  20s,'' 2011.

\bibitem{beyer2022knowledge}
L.~Beyer, X.~Zhai, A.~Royer, L.~Markeeva, R.~Anil, and A.~Kolesnikov,
  ``Knowledge distillation: A good teacher is patient and consistent,'' in {\em
  CVPR}, pp.~10925--10934, 2022.

\bibitem{lee2018snip}
N.~Lee, T.~Ajanthan, and P.~Torr, ``{SNIP}: {SINGLE}-{SHOT} {NETWORK} {PRUNING}
  {BASED} {ON} {CONNECTION} {SENSITIVITY},'' in {\em ICLR}, 2019.

\bibitem{lin2022fairgrape}
X.~Lin, S.~Kim, and J.~Joo, ``Fairgrape: Fairness-aware gradient pruning method
  for face attribute classification,'' in {\em ECCV}, pp.~414--432, Springer,
  2022.

\bibitem{liu2021discrimination}
J.~Liu, B.~Zhuang, Z.~Zhuang, Y.~Guo, J.~Huang, J.~Zhu, and M.~Tan,
  ``Discrimination-aware network pruning for deep model compression,'' {\em
  {IEEE} Trans. Pattern Anal. Mach. Intell.}, vol.~44, no.~8, pp.~4035--4051,
  2021.

\bibitem{chen2018mobilefacenets}
S.~Chen, Y.~Liu, X.~Gao, and Z.~Han, ``Mobilefacenets: Efficient cnns for
  accurate real-time face verification on mobile devices,'' in {\em {CCBR}},
  vol.~10996 of {\em Lecture Notes in Computer Science}, pp.~428--438,
  Springer, 2018.

\bibitem{IJCB2023JanColor}
J.~N. Kolf, J.~Elliesen, F.~Boutros, and N.~Damer, ``How colorful should faces
  be? harmonizing color and model quantization for resource-restricted face
  recognition,'' in {\em IJCB}, pp.~1--10, {IEEE} (to appear), 2023.

\bibitem{howard2017mobilenets}
A.~G. Howard, M.~Zhu, B.~Chen, D.~Kalenichenko, W.~Wang, T.~Weyand,
  M.~Andreetto, and H.~Adam, ``Mobilenets: Efficient convolutional neural
  networks for mobile vision applications,'' {\em arXiv preprint
  arXiv:1704.04861}, 2017.

\bibitem{vaswani2017attention}
A.~Vaswani, N.~Shazeer, N.~Parmar, J.~Uszkoreit, L.~Jones, A.~N. Gomez,
  {\L}.~Kaiser, and I.~Polosukhin, ``Attention is all you need,'' {\em
  NEURIPS}, vol.~30, 2017.

\bibitem{dosovitskiy2020image}
A.~Dosovitskiy, L.~Beyer, A.~Kolesnikov, D.~Weissenborn, X.~Zhai,
  T.~Unterthiner, M.~Dehghani, M.~Minderer, G.~Heigold, S.~Gelly, J.~Uszkoreit,
  and N.~Houlsby, ``An image is worth 16x16 words: Transformers for image
  recognition at scale,'' in {\em ICLR}, 2021.

\bibitem{ji2022neural}
M.~Ji, G.~Peng, S.~Li, F.~Cheng, Z.~Chen, Z.~Li, and H.~Du, ``A neural network
  compression method based on knowledge-distillation and parameter quantization
  for the bearing fault diagnosis,'' {\em Applied Soft Computing}, vol.~127,
  p.~109331, 2022.

\bibitem{zoph2016neural}
B.~Zoph and Q.~Le, ``Neural architecture search with reinforcement learning,''
  in {\em ICLR}, 2017.

\bibitem{he2016deep}
K.~He, X.~Zhang, S.~Ren, and J.~Sun, ``Deep residual learning for image
  recognition,'' in {\em CVPR}, pp.~770--778, 2016.

\bibitem{li2019graph}
D.~Li, G.~Wen, X.~Li, and X.~Cai, ``Graph-based dynamic ensemble pruning for
  facial expression recognition,'' {\em Applied Intelligence}, vol.~49,
  pp.~3188--3206, 2019.

\bibitem{xu2023probabilistic}
J.~Xu, S.~Li, A.~Deng, M.~Xiong, J.~Wu, J.~Wu, S.~Ding, and B.~Hooi,
  ``Probabilistic knowledge distillation of face ensembles,'' in {\em CVPR},
  pp.~3489--3498, 2023.

\bibitem{parkhi2015deep}
O.~Parkhi, A.~Vedaldi, and A.~Zisserman, ``Deep face recognition,'' in {\em
  BMVC}, British Machine Vision Association, 2015.

\bibitem{alonso2023squeezerfacenet}
F.~Alonso{-}Fernandez, K.~Hernandez{-}Diaz, J.~M.~B. Rubio, and J.~Big{\"{u}}n,
  ``Squeezerfacenet: Reducing a small face recognition {CNN} even more via
  filter pruning,'' in {\em {IWAIPR}}, vol.~14335 of {\em Lecture Notes in
  Computer Science}, pp.~349--361, Springer, 2023.

\bibitem{frankle2018lottery}
J.~Frankle and M.~Carbin, ``The lottery ticket hypothesis: Finding sparse,
  trainable neural networks,'' in {\em ICLR}, 2019.

\bibitem{mirzadeh2020improved}
S.~I. Mirzadeh, M.~Farajtabar, A.~Li, N.~Levine, A.~Matsukawa, and
  H.~Ghasemzadeh, ``Improved knowledge distillation via teacher assistant,'' in
  {\em AAAI}, vol.~34, pp.~5191--5198, 2020.

\bibitem{blakeney2021simon}
C.~Blakeney, N.~Huish, Y.~Yan, and Z.~Zong, ``Simon says: Evaluating and
  mitigating bias in pruned neural networks with knowledge distillation,'' {\em
  arXiv preprint arXiv:2106.07849}, 2021.

\bibitem{robinson2020face}
J.~P. Robinson, G.~Livitz, Y.~Henon, C.~Qin, Y.~Fu, and S.~Timoner, ``Face
  recognition: too bias, or not too bias?,'' in {\em CVPRW}, pp.~0--1, 2020.

\bibitem{DBLP:conf/bmvc/AlbieroB20}
V.~Albiero and K.~W. Bowyer, ``Is face recognition sexist? no, gendered
  hairstyles and biology are,'' in {\em {BMVC}}, 2020.

\bibitem{DBLP:conf/icb/AlbieroZB20}
V.~Albiero, K.~Zhang, and K.~W. Bowyer, ``How does gender balance in training
  data affect face recognition accuracy?,'' in {\em {IJCB}}, pp.~1--10, {IEEE},
  2020.

\bibitem{DBLP:conf/icb/FuD22}
B.~Fu and N.~Damer, ``Towards explaining demographic bias through the eyes of
  face recognition models,'' in {\em {IJCB}}, pp.~1--10, {IEEE}, 2022.

\bibitem{DBLP:conf/icb/DebN018}
D.~Deb, N.~Nain, and A.~K. Jain, ``Longitudinal study of child face
  recognition,'' in {\em {ICB}}, pp.~225--232, {IEEE}, 2018.

\bibitem{DBLP:journals/pami/HuangLLT20}
C.~Huang, Y.~Li, C.~C. Loy, and X.~Tang, ``Deep imbalanced learning for face
  recognition and attribute prediction,'' {\em {IEEE} Trans. Pattern Anal.
  Mach. Intell.}, vol.~42, no.~11, pp.~2781--2794, 2020.

\bibitem{terhorst2020compr}
P.~Terhörst, J.~N. Kolf, M.~Huber, F.~Kirchbuchner, N.~Damer, A.~M. Moreno,
  J.~Fierrez, and A.~Kuijper, ``A comprehensive study on face recognition
  biases beyond demographics,'' {\em IEEE Transactions on Technology and
  Society}, vol.~3, no.~1, pp.~16--30, 2022.

\bibitem{fang2024fairpad}
M.~Fang, W.~Yang, A.~Kuijper, V.~Struc, and N.~Damer, ``Fairness in face
  presentation attack detection,'' {\em Pattern Recognition}, vol.~147,
  p.~110002, 2024.

\bibitem{DBLP:conf/eusipco/FangDKK20}
M.~Fang, N.~Damer, F.~Kirchbuchner, and A.~Kuijper, ``Demographic bias in
  presentation attack detection of iris recognition systems,'' in {\em
  {EUSIPCO}}, pp.~835--839, {IEEE}, 2020.

\bibitem{DBLP:conf/icb/TerhorstKDKK20}
P.~Terh{\"{o}}rst, J.~N. Kolf, N.~Damer, F.~Kirchbuchner, and A.~Kuijper,
  ``Face quality estimation and its correlation to demographic and
  non-demographic bias in face recognition,'' in {\em {IJCB}}, pp.~1--11,
  {IEEE}, 2020.

\bibitem{DBLP:conf/fgr/MittalTMVS23}
S.~Mittal, K.~Thakral, P.~Majumdar, M.~Vatsa, and R.~Singh, ``Are face
  detection models biased?,'' in {\em {FG}}, pp.~1--7, {IEEE}, 2023.

\bibitem{DBLP:conf/eusipco/HuberFBD23}
M.~Huber, M.~Fang, F.~Boutros, and N.~Damer, ``Are explainability tools gender
  biased? {A} case study on face presentation attack detection,'' in {\em
  {EUSIPCO}}, pp.~945--949, {IEEE}, 2023.

\bibitem{wang2019racial}
M.~Wang, W.~Deng, J.~Hu, X.~Tao, and Y.~Huang, ``Racial faces in the wild:
  Reducing racial bias by information maximization adaptation network,'' in
  {\em ICCV}, pp.~692--702, 2019.

\bibitem{karkkainen2021fairface}
K.~Karkkainen and J.~Joo, ``Fairface: Face attribute dataset for balanced race,
  gender, and age for bias measurement and mitigation,'' in {\em WACV},
  pp.~1548--1558, 2021.

\bibitem{wang2021meta}
M.~Wang, Y.~Zhang, and W.~Deng, ``Meta balanced network for fair face
  recognition,'' {\em {IEEE} Trans. Pattern Anal. Mach. Intell.}, vol.~44,
  no.~11, pp.~8433--8448, 2021.

\bibitem{xu2020investigating}
T.~Xu, J.~White, S.~Kalkan, and H.~Gunes, ``Investigating bias and fairness in
  facial expression recognition,'' in {\em ECCVW}, pp.~506--523, Springer,
  2020.

\bibitem{stoychev2022effect}
S.~Stoychev and H.~Gunes, ``The effect of model compression on fairness in
  facial expression recognition,'' vol.~13646, pp.~121--138, 2022.

\bibitem{DBLP:conf/iccvw/LiuZSYL21}
B.~Liu, S.~Zhang, G.~Song, H.~You, and Y.~Liu, ``Rectifying the data bias in
  knowledge distillation,'' in {\em {ICCVW}}, pp.~1477--1486, {IEEE}, 2021.

\bibitem{DBLP:conf/cvpr/IofinovaPA23}
E.~Iofinova, A.~Peste, and D.~Alistarh, ``Bias in pruned vision models:
  In-depth analysis and countermeasures,'' in {\em {CVPR}}, pp.~24364--24373,
  {IEEE}, 2023.

\bibitem{ahn-etal-2022-knowledge}
J.~Ahn, H.~Lee, J.~Kim, and A.~Oh, ``Why knowledge distillation amplifies
  gender bias and how to mitigate from the perspective of {D}istil{BERT},'' in
  {\em Proceedings of the 4th Workshop on Gender Bias in Natural Language
  Processing (GeBNLP)}, pp.~266--272, Association for Computational
  Linguistics, 2022.

\bibitem{goncalves-strubell-2023-understanding}
G.~Gon{\c{c}}alves and E.~Strubell, ``Understanding the effect of model
  compression on social bias in large language models,'' in {\em Proceedings of
  the 2023 Conference on Empirical Methods in Natural Language Processing},
  pp.~2663--2675, Association for Computational Linguistics, 2023.

\bibitem{paganini2020prune}
M.~Paganini, ``Prune responsibly,'' {\em arXiv preprint arXiv:2009.09936},
  2020.

\end{thebibliography}


\end{document}